\documentclass[10pt]{article}
\usepackage[preprint]{tmlr}
\renewcommand{\headrulewidth}{0pt}
\usepackage{microtype}
\usepackage{graphicx}
\usepackage{subcaption}
\usepackage{booktabs}

\usepackage{amsmath}
\usepackage{amssymb}
\usepackage{mathtools}
\usepackage{amsthm}

\usepackage{url}
\usepackage{enumitem}
\usepackage{algorithm}
\usepackage{pythonhighlight}

\usepackage{needspace}

\makeatletter
\@ifundefined{lststy@pythonhighlight-style$}
    {\lstdefinestyle{pythonhighlight-style}{style=mypython}}{}
\makeatother

\usepackage[hidelinks]{hyperref}
\hypersetup{
  pdftitle={Load Balancing Neurons: Firing-Rate Control Sustains Plasticity in Continual Learning},
  pdfauthor={Jan Robine, Marc H\"oftmann, Stefan Harmeling}}
\usepackage[capitalize,noabbrev]{cleveref}

\Crefname{equation}{Equation}{Equations}
\Crefname{figure}{Figure}{Figures}
\Crefname{algorithm}{Algorithm}{Algorithms}
\crefname{algorithm}{Algorithm}{Algorithms}
\Crefname{tabular}{Table}{Tables}
\Crefname{section}{Section}{Sections}

\definecolor{codeblue}{RGB}{0,110,174}
\colorlet{literatecolour}{black}
\lstdefinestyle{papercode}{style=pythonhighlight-style,
  keywordstyle=\color{codeblue}\bfseries,
  emphstyle=\color{codeblue}\bfseries,
  emphstyle=[2]\color{codeblue}\bfseries,
  emphstyle=[3]\color{black},
  emphstyle=[4]\color{black},
  emphstyle=[5]\color{black},
  emphstyle=[6]\color{codeblue}\bfseries,
  emphstyle=[7]\color{codeblue}\bfseries,
  stringstyle=\color{black}}

\lstnewenvironment{mypython}[1][]{\lstset{style=papercode, frame=none, morekeywords={with}, #1}}{}

\usepackage{amsmath,amsfonts,bm}

\def\eqref#1{equation~\ref{#1}}

\def\1{\bm{1}}

\def\rvs{{\mathbf{s}}}

\def\rvx{{\mathbf{x}}}

\DeclareMathAlphabet{\mathsfit}{\encodingdefault}{\sfdefault}{m}{sl}
\SetMathAlphabet{\mathsfit}{bold}{\encodingdefault}{\sfdefault}{bx}{n}

\newcommand{\E}{\mathbb{E}}

\newcommand{\R}{\mathbb{R}}

\newcommand{\revision}[1]{#1}

\def\firingrate{{\rho}}
\def\avgfiringrate{{\bar{\rho}}}
\def\minfiringrate{{\firingrate_\text{min}}}
\def\maxfiringrate{{\firingrate_\text{max}}}
\def\avgmomentum{{\tau}}
\def\E{\mathbb{E}}
\def\R{\mathbb{R}}
\DeclareMathOperator{\sg}{sg}

\title{Load Balancing Neurons: Firing-Rate Control\\Sustains Plasticity in Continual Learning}

\author{\name Jan Robine \email jan.robine@tu-dortmund.de \\
      \addr Department of Computer Science\\
      TU Dortmund University\\
      Lamarr Institute for Machine Learning and Artificial Intelligence
      \AND
      \name Marc H\"oftmann \email marc.hoeftmann@tu-dortmund.de \\
      \addr Department of Computer Science\\
      TU Dortmund University\\
      Lamarr Institute for Machine Learning and Artificial Intelligence
      \AND
      \name Stefan Harmeling \email stefan.harmeling@tu-dortmund.de \\
      \addr Department of Computer Science\\
      TU Dortmund University\\
      Lamarr Institute for Machine Learning and Artificial Intelligence}

\begin{document}

\maketitle

\begin{abstract}
Neural networks in continual learning often lose plasticity: some neurons become inactive, while others fire almost constantly. This limits adaptation to shifting data and wastes capacity. Prior work mitigates this by periodically reinitializing low-utility units, but such resets can destroy previously learned features and do not proactively prevent low utility. We study a simple diagnostic measure: the firing rate of ReLU units, defined as the fraction of positive pre-activations. Low rates identify dead units, while very high rates indicate linearized, always-on units. Based on this view, we introduce a lightweight load-balancing mechanism that adjusts per-neuron thresholds to keep firing rates within a target range. Across continual ImageNet and class-incremental CIFAR-100, firing-rate distributions explain differences in plasticity across a range of approaches, including well-known techniques such as L2 regularization and non-affine normalization, and load balancing matches or exceeds reset-based methods on both benchmarks.
The same mechanism carries over to reinforcement learning, improving Rainbow DQN on Atari over a LayerNorm baseline.
\end{abstract}

\section{Introduction}
\label{sec:introduction}

Deep learning models have achieved state-of-the-art performance on a broad range of tasks when trained on fixed datasets \citep{he2016deep, liu2022convnet}.
However, performance often degrades in continual learning settings, where the data distribution evolves over time \citep{khetarpal2022towards, lee2023plastic, abbas2023loss}.
Simply continuing training on new data often leads to optimization instabilities and catastrophic forgetting, and the mechanisms behind these failures are not yet fully understood \citep{lyle2024disentangling}.
A common remedy is to retrain a new model from scratch, which restores accuracy but is often impractical due to computational cost.

Ideally, a continual learning system would strike a balance between \emph{plasticity}, the ability to adapt to new data, and \emph{stability}, the ability to preserve what it has already learned \citep{carpenter1987self,elsayed2024addressing}.
Declining stability is known as \emph{catastrophic forgetting} \citep{mccloskey1989catastrophic}, while declining plasticity is commonly called \emph{plasticity loss}.
Overcoming these challenges would allow models to evolve as data changes over time, mimicking the way humans and animals continuously learn without retraining from scratch.

Plasticity loss has a simple signature at the level of individual units.
In a ReLU network, a unit that has stopped firing contributes neither activation nor gradient, while a unit that fires on almost every input has become an affine pass-through; in both cases its threshold no longer separates anything.
We make this observable by tracking one number per unit --- its \emph{firing rate}, the fraction of inputs for which it is active --- and find that the distribution of these rates tracks plasticity across methods, benchmarks and architectures.
It is also directly controllable: a per-neuron offset, updated by a controller rather than by gradient descent, holds firing rates inside a target range and improves accuracy over reset-based methods on both supervised benchmarks and over a LayerNorm baseline on Atari.
The same lens explains why several unrelated known remedies work, and suggests architectural changes that induce balanced rates with no explicit control at all.

\needspace{5\baselineskip}
\textbf{Contributions:}
\begin{itemize}[itemsep=1pt,topsep=0pt]
\item We introduce firing-rate analysis as a simple, interpretable diagnostic tool for neuron utilization and plasticity loss in continual learning, offering a unified view across tasks and architectures.

\item We analyze and compare the firing-rate distributions of existing methods, including L2 regularization and non-affine normalization, and show how these patterns relate to plasticity loss.

\item We propose a load-balancing mechanism that dynamically adjusts neuron activity by shifting activation inputs. It keeps firing rates balanced and outperforms previous baselines.
\end{itemize}

\section{Background}
\label{sec:background}

\revision{
\paragraph{Plasticity loss.}
The phenomenon of plasticity loss has been observed in both supervised learning \citep{ash2020warm,berariu2021study,dohare2024loss} and reinforcement learning \citep{lyle2022understanding,nikishin2022primacy,dohare2024loss}.
Several approaches have been proposed to improve plasticity.
For instance, injecting noise during training \citep{ash2020warm},
periodically resetting the last layers \citep{nikishin2022primacy},
expanding the network during training \citep{nikishin2023deep},
using the Concatenated ReLU activation function \citep{abbas2023loss},
reinitializing low-utility neurons \citep{sokar2023dormant,dohare2024loss},
self-normalized neuron resets based on inactivity statistics \citep{farias2025selfnormalized},
perturbing gradients of low-utility neurons \citep{elsayed2024addressing},
and regularizing weights toward their initial values \citep{kumar2024maintaining}.
Some studies combine multiple techniques, for example, \citet{lee2023plastic} combines layer resets, Concatenated ReLUs, LayerNorm \citep{ba2016layer}, and sharpness-aware minimization \citep{foret2021sharpness}.

Many of these interventions restore plasticity at the cost of destroying learned information: they reset neurons or entire layers rather than preventing collapse in the first place.
In contrast, we ask how to sustain plasticity proactively through minimal architectural changes.

\paragraph{Measuring neuron utility.}
Many of the above methods rely on a measure of neuron utility.
For example, \citet{dohare2024loss} define a neuron's \emph{contribution utility} as the magnitude of its activation multiplied by the magnitudes of all outgoing weights, tracked as a moving average during training.
\citet{sokar2023dormant} define the \emph{dormancy score} as the neuron's activation magnitude, normalized by the magnitudes of other neurons in the same layer and averaged over the dataset;
a neuron is considered \emph{dormant} if the dormancy score falls below a threshold.
\citet{elsayed2024addressing} define \emph{weight utility} as the change in loss after zeroing a weight, which yields a global ranking over all weights but is more expensive to compute.
Self-Normalized Resets instead applies a self-normalized percentile test to inter-firing times, using inactivity statistics rather than magnitudes to decide when to reset \citep{farias2025selfnormalized}.

In contrast, we introduce a firing-rate diagnostic that reveals the \emph{distribution} of neuron utilization, across layers and over time, rather than collapsing activity into a single summary statistic.
}

\paragraph{Continual learning benchmarks.}
Many benchmarks have been proposed to evaluate continual learning, but there is no single standard, and new variants are often introduced, making comparisons difficult.
In this work, we adopt the benchmarking setup of \citet{dohare2024loss} to assess our method.
We therefore focus on the \emph{continual ImageNet} and \emph{class-incremental \mbox{CIFAR-100}} benchmarks, which they used to evaluate continual backpropagation (CBP).

\section{Plasticity through the lens of firing rates}
\label{sec:firing-rate}

\revision{
Plasticity loss often reflects neuron-level collapse in ReLU networks, where some neurons stop firing and others become always active.
Plasticity, in turn, depends on whether a neuron spends time on both sides of its threshold.
We therefore use the \emph{firing rate}, the fraction of pre-activations for which a neuron is active, as a diagnostic tool of neuron use.
Tracking firing-rate distributions reveals training anomalies, and keeping them in a balanced range correlates with improved adaptation under distribution shift, as we will demonstrate in the experimental section.

\subsection{Plasticity as the effective number of activation regions}

To make this intuition more precise, we now view the network through the lens of its piecewise-affine structure and its binary activation patterns.
Many standard network designs, like multilayer perceptrons and convolutional nets using ReLU or similar functions, implement a \emph{continuous piecewise-affine (CPWA) map} that partitions the input space into regions on which the network is affine \citep{montufar2014linear,serra2018bounding,goujon2024linear}.
Each such region is determined by which hidden units are active or inactive, and there is a one-to-one correspondence between these activation patterns and the affine regions induced by the network \citep{zhang2020empirical}.
This viewpoint lets us characterize the network’s functional complexity in terms of how finely these regions carve up the input space \citep{montufar2014linear,serra2018bounding,zhang2020empirical}; extensions to non-CPWA architectures are discussed in \cref{sec:further-theory}.

Consider a network with parameters $\theta$ and a total of $M$ hidden units.
For any input ${x \in \mathcal{X}}$, let $z_i(x)$ denote the pre-activation of unit ${i \in \{1,\dots,M\}}$.
We define the corresponding binary gate
\begin{equation}
    S_i(x) \coloneqq [z_i(x) \geq 0],
\end{equation}
which indicates whether unit $i$ is active.
The \emph{global activation pattern} is the vector of all gates:
\begin{equation}
    S(x) \coloneqq \bigl(S_1(x), \ldots, S_M(x)\bigr) \;\in\; \{0,1\}^M.
\end{equation}
Each unique pattern ${s \in \{0,1\}^M}$ defines a region ${\mathcal{R}(s) \coloneqq \{ x \mid S(x) = s \}}$ where the network is affine.
All non-empty regions together form a partition $\mathcal{P}$ of the input space.

\paragraph{Effective capacity and plasticity.}
The expressive power of a CPWA network is fundamentally determined by its partition $\mathcal{P}$ \citep{montufar2014linear,serra2018bounding,takai2021linear}.
Since the network is an affine map on each region ${R \in \mathcal{P}}$, it cannot capture non-affine variations within a single region.
Thus, to represent a nonlinear target function, the partition must separate inputs that cannot be well-approximated by the same affine function into different regions.

A naive way to quantify this capacity is to count the total number of regions $|\mathcal{P}|$, but this treats all regions equally, regardless of how often they are visited by the data.
To obtain a data-dependent notion, let ${\rvx \sim P_\rvx}$ and let ${\rvs \coloneqq S(\rvx)}$ be the random activation pattern ($\rvx$, $\rvs$ are random variables; $x$, $s$ their values).
We then measure the effective number of regions using the \emph{perplexity} $2^{H(\rvs)}$ of the activation patterns.
Standard entropy bounds yield ${2^{H(\rvs)} \leq \lvert \mathrm{supp}(\rvs) \rvert \leq \lvert \mathcal{P} \rvert}$, where $\mathrm{supp}(\rvs)$ is the support of $\rvs$, so the perplexity provides a smooth, data-dependent lower bound on the number of regions that carry non-negligible probability mass under $P_\rvx$.
Related work has used entropy and perplexity of activation patterns to quantify how richly a network distinguishes inputs or classes \citep{davel2020dnns,hartmann2021studying}.

Since the mapping ${x \mapsto S(x)}$ is deterministic, ${H(\rvs \mid \rvx) = 0}$, so ${I(\rvx; \rvs) = H(\rvs) - H(\rvs \mid \rvx) = H(\rvs)}$.
Maximizing this lower bound is therefore equivalent to maximizing the mutual information between the input and its activation pattern, forcing the discrete code to retain maximal information about the input (the \emph{InfoMax principle}; \citealp{linsker1988self}).
Closely related, mutual-information objectives between activation codes and inputs have been used to explicitly increase the nonlinear expressivity of encoders \citep{park2021unsupervised}.
In a continual learning setting with a time-varying distribution $P_{\rvx_t}$, maximizing $2^{H(\rvs_t)}$ continually pushes up the lower bound on $|\mathcal{P}|$.
This is promoting \emph{plasticity}, since it encourages the network to keep many different affine regions actively used, so it can keep telling apart new and changing input patterns.

\subsection{A tractable proxy: per-neuron firing rates}
\label{sec:entropy-objective}

The global entropy $H(\rvs)$ is bounded above by the sum of the entropies of the individual units.
The gap is the \emph{total correlation (TC)} \citep{watanabe1960information}, which measures the redundancy (statistical dependence) among units:
\begin{equation}
    \label{eq:tc_decomp}
    H(\rvs) = \sum_{i=1}^M H(\rvs_i) - TC(\rvs).
\end{equation}
Optimizing the joint entropy $H(\rvs)$ directly is computationally challenging due to the difficulty of estimating high-dimensional dependencies (the $TC$ term).
However, the first term in \cref{eq:tc_decomp}, $\sum_{i=1}^M H(\rvs_i)$, depends only on marginal probabilities and is efficient to compute.
Since each gate $\rvs_i$ is a Bernoulli variable, its entropy $H(\rvs_i)$ is maximized when its probability of being active is exactly $0.5$.
Similar observations are used in binarized activation entropy regularizers that encourage individual units to be active about half the time while keeping them weakly correlated, in order to increase the joint entropy of their activation patterns \citep{cao2018improving}.
Therefore, maximizing the sum of marginal entropies encourages every unit to fire for half of the inputs, avoiding the ``dead unit'' problem and maximizing the information capacity of individual neurons.

We propose to maximize this sum as a simple, tractable stand-in for global plasticity.
In principle, this objective can still leave redundancy between units (i.e., when $TC(\rvs)$ is non-zero), but we observe that making each unit active about half the time will usually also increase the joint entropy:
In \cref{sec:loadbalancing} we check this assumption empirically and find that the total correlation stays under control while the marginal entropies grow, so the extra unit-wise capacity corresponds to a real increase in the effective number of affine regions.
A complementary view is in terms of the average Hamming distance between activation codes, which we discuss further in \cref{sec:further-theory}.

\paragraph{Firing rate.}
The firing-rate perspective provides a practical way to monitor and control the marginal entropies of the gates $\rvs_i$.
For each unit $i$, the gate $\rvs_i$ is a Bernoulli random variable under ${\rvx \sim P_\rvx}$, with activation probability ${\Pr(\rvs_i = 1) = \Pr(z_i(\rvx) \ge 0)}$.
In practice, we estimate this probability from a mini-batch ${x_1, \ldots, x_n}$ via the \emph{instantaneous firing rate}
\begin{equation}
  \label{eq:firing-rate}
  \firingrate_i \coloneqq \frac{1}{n} \sum_{j=1}^n [ z_i(x_j) \ge 0 ],
\end{equation}
which is the empirical mean of the gate $\rvs_i$ over the batch.
To smooth batch noise and remain responsive to a changing data distribution, we track an \emph{exponentially averaged firing rate} with decay ${\avgmomentum \in [0, 1]}$, denoted $\avgfiringrate_i$, to capture the longer-term trend:
\begin{equation}
  \label{eq:avg-firing-rate}
  \avgfiringrate_i \leftarrow \avgmomentum \avgfiringrate_i + (1 - \avgmomentum) \firingrate_i.
\end{equation}

For ReLU units, the pre-activation $z_i$ induces two regimes: an \emph{off} regime ${z_i < 0}$ (zero output and zero slope) and an \emph{on} regime ${z_i \geq 0}$ (positive output and unit slope).
The firing rate $\avgfiringrate_i$ thus directly reflects how often unit $i$ is in each regime.
Near ${\avgfiringrate_i = 0}$ the neuron is \emph{dead} or \emph{dormant} and contributes neither activation nor gradient.
Near ${\avgfiringrate_i = 1}$ it behaves as an almost linear pass-through with gradients still flowing, but adding no curvature.
In both extremes, the entropy of the gate $\rvs_i$ is close to zero, and the unit contributes little to the global entropy $H(\rvs)$, reducing effective capacity and plasticity.
This view motivates using firing rates both as a diagnostic for plasticity and as a control signal for the load-balancing mechanism introduced later.

}

\subsection{Firing-rate dynamics in continual benchmarks}

In this subsection, we study firing-rate behavior on the continual ImageNet and class-incremental CIFAR-100 benchmarks.
These settings build on widely used datasets, are tailored to CNN and ResNet architectures that provide a stronger test of plasticity than simpler MLP models, and are straightforward to interpret.
Together, they offer a useful testbed for examining how neuron firing rates evolve under distribution shift.

\begin{figure}[t]
    \centerline{%
        \begin{minipage}[c]{0.5\textwidth}\centering%
            \includegraphics{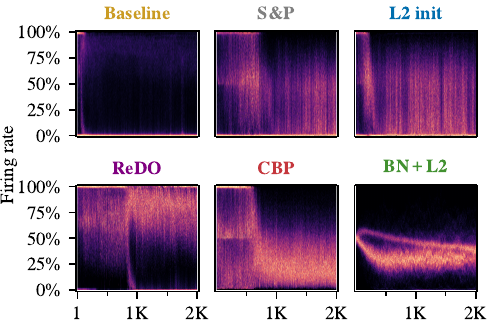}%
        \end{minipage}%
        \begin{minipage}[c]{0.5\textwidth}\centering%
            \includegraphics{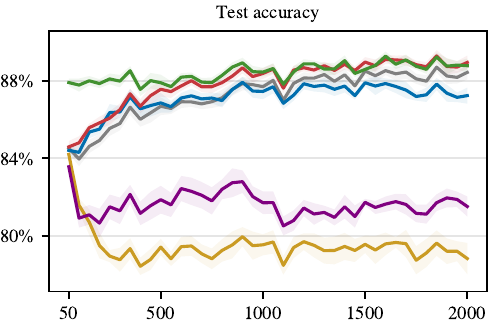}%
        \end{minipage}%
    }
    \caption{\textbf{Continual ImageNet:}
    Evolution of firing-rate distributions (left) and test accuracy with error bands (right), averaged over 10 runs.
    Standard training collapses neuron activity toward $0\%$, while BatchNorm, L2 regularization and CBP keep firing more balanced.
    The temporal alignment between balanced firing rates and rising accuracy is what connects the two; per-method findings are in the main text.
    }
    \label{fig:imagenet-firing-rates}
\end{figure}

\paragraph{Continual ImageNet: firing-rates and accuracy.}

In the continual ImageNet benchmark \citep{dohare2024loss}, the model is trained on a sequence of binary classification tasks.
For each task it learns to discriminate two randomly sampled classes from ImageNet \citep{deng2009imagenet} 
before moving on to a new pair.
Because all tasks have comparable difficulty, accuracy degradation across tasks indicates reduced plasticity \citep{dohare2024loss}.
Following \citet{dohare2024loss}, images are downsampled to $32\times32$ resolution, which preserves ImageNet's class diversity while making long sequential runs computationally feasible for thousands of tasks.
This setup is substantially more demanding than MNIST-scale benchmarks.
We use a compact CNN with ReLU activations (see \cref{sec:training-details} for training details).

Firing-rate distributions and test accuracies of several baselines are shown in \cref{fig:imagenet-firing-rates}.
Hyperparameter details are provided in \cref{sec:training-details}.
Accuracy curves report the mean over 10 runs, with shaded bands denoting $\pm 1$ standard error of the mean.
Because each task is a randomly sampled pair of classes, task difficulty varies considerably, and this task-sampling variance dominates the spread between runs of any single method.
It is, however, \emph{shared}: every method is evaluated on the same 10 task sequences, so it cancels when methods are compared run by run.
We therefore report differences between methods as paired per-run differences (\cref{tab:imagenet-accuracy}), which resolve a comparison far more precisely than either method's mean does on its own.
Absolute accuracies for all baselines and tasks are listed in \cref{sec:accuracy-tables}.

Throughout this paper, firing-rate distributions are visualized as two-dimensional heatmaps:
the horizontal axis denotes training progress (number of tasks), the vertical axis spans firing rates from 0 to 100\%, and each column shows a histogram over all neurons in the network.
Brighter regions indicate many neurons with that firing rate, whereas darker regions indicate fewer.
These plots reveal how activity patterns evolve over time.

Key findings:
\begin{enumerate}[topsep=0pt,itemsep=0pt]
\item \textbf{Baseline.}
Accuracy degrades rapidly over tasks.
Firing rates collapse toward ${\approx 0\%}$ for most neurons, aligning with the drop in accuracy.

\item \textbf{S\&P.} Shrink-and-perturb combines L2 regularization with small parameter noise \citep{ash2020warm}. It increases the fraction of active neurons (up to ${\approx 75\%}$) and substantially improves accuracy, though many neurons remain near ${\approx 0\%}$.

\item \textbf{L2 init.} L2 init regularizes weights toward their initialization \citep{kumar2024maintaining}. Like S\&P, it yields higher activity (up to ${\approx 80\%}$) and higher accuracy, while a subset still collapses toward ${\approx 0\%}$.

\item \textbf{ReDO.} Recycling dormant neurons resets neurons that fall below a dormancy threshold \citep{sokar2023dormant}. It largely eliminates the ${\approx 0\%}$ regime, but leads to widespread saturation early (${\approx 100\%}$) and sustained high firing (${\approx 75{-}95\%}$), consistent with linearized ReLUs, and with lower accuracy than the other interventions.

\item \textbf{CBP.}
Continual backpropagation resets ``underutilized'' neurons using an activation-magnitude criterion \citep{dohare2024loss}. It largely prevents collapse to ${\approx 0\%}$, but early saturation near ${\approx 100\%}$ again suggests linearized ReLUs.
After ${\approx 500}$ tasks, the distribution recenters and broadens, coinciding with improved accuracy.

\item \textbf{BatchNorm\,+\,L2.}
Combining BatchNorm \citep{ioffe2015batch} with L2 regularization already prevents extreme firing rates. Both accuracy and plasticity stay high.
We abbreviate this configuration BN$\,+\,$L2 in figures and tables, and likewise write LN for LayerNorm.

\end{enumerate}

\begin{figure}[t]
    \centerline{%
        \begin{minipage}[c]{0.5\textwidth}\centering%
            \includegraphics{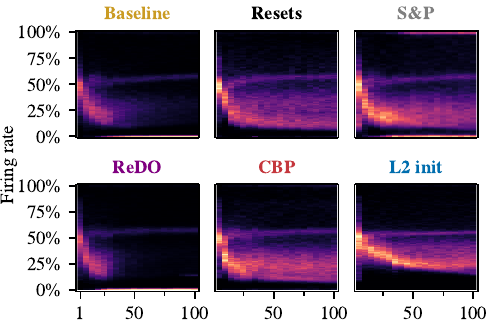}%
        \end{minipage}%
        \begin{minipage}[c]{0.5\textwidth}\centering%
            \includegraphics{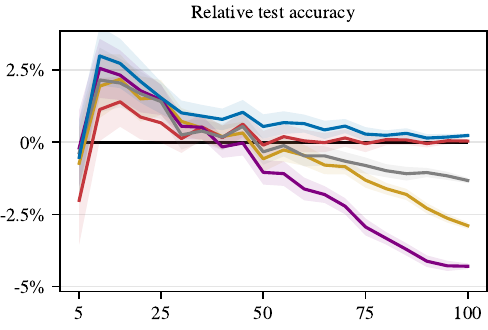}%
        \end{minipage}%
    }
    \caption{\textbf{Class-incremental CIFAR-100:}
    Firing-rate distributions (left) and test accuracy with error bands (right), averaged over 15 runs.
    Accuracy is relative to the from-scratch \emph{Resets} baseline, which is the zero line.
    Continual training drives neurons gradually inactive, while resets and CBP keep firing patterns from collapsing, and accuracy follows those trends; per-method findings are in the main text.
    }
    \label{fig:cifar100-firing-rates}
\end{figure}

\paragraph{Class-incremental CIFAR-100: firing-rates and accuracy.}

In the class-incremental CIFAR-100 benchmark, the model begins by learning to classify a small set of randomly sampled classes from CIFAR-100 \citep{krizhevsky2009learning}.
Additional classes are gradually introduced, and the model must classify all classes encountered so far until the full set of 100 is included.
Because the classification problem grows harder as more classes accumulate, performance can decline even without plasticity loss \citep{dohare2024loss}.
To disentangle this effect, we compare continual training with a model trained from scratch on the same subset.
If the continually trained model performs worse than a model trained from scratch, this indicates plasticity loss, whereas superior performance signals knowledge transfer \citep{dohare2024loss}.
We use a ResNet-18 with ReLU activations (see \cref{sec:training-details} for training details).

Firing-rate distributions and test accuracies of several baselines are shown in \cref{fig:cifar100-firing-rates}.
Results are averaged over 15 runs, with shaded bands denoting $\pm 1$ standard error of the mean.

Throughout this paper, class-incremental CIFAR-100 accuracy is plotted \emph{relative} to the from-scratch \emph{Resets} baseline, following \citet{dohare2024loss}.
\emph{Resets} is therefore the zero line in every right-hand panel and carries no curve of its own: a curve above zero indicates knowledge transfer, and one below it indicates plasticity loss.
Absolute accuracies for all methods are listed in \cref{sec:accuracy-tables}.

Key findings:
\begin{enumerate}[topsep=0pt,itemsep=0pt]
\item \textbf{Baseline.}
Neurons gradually become inactive, though less severely than continual ImageNet.
Many units reach $\approx0\%$ firing, limiting the capacity for new tasks.

\item \textbf{Resets.}
Reinitializing the full network after each task prevents collapse and preserves adaptability.
As tasks become harder, the firing-rate distribution shifts accordingly.

\item \textbf{S\&P.}
Firing-rate distributions move toward the \emph{Resets} regime, but many neurons still cluster near ${\approx 0\%}$ (dead) and ${\approx 100\%}$ (effectively linearized). This imbalance corresponds to lower accuracy than \emph{Resets}.

\item \textbf{ReDO.}
Like the baseline, most neurons become inactive; accuracy is even lower than the baseline.

\item \textbf{CBP.}
Firing-rate distributions closely match \emph{Resets}, and test accuracy is comparable.
Most units lie around 10--30\%, with some up to ${\approx 60\%}$, indicating sustained plasticity.

\item \textbf{L2 init.}
Firing rates concentrate even more tightly around ${\approx 50\%}$ than with CBP. It achieves the highest accuracy among these baselines.
\end{enumerate}

\paragraph{Summary of findings.}
In summary, firing-rate analysis provides a straightforward lens on plasticity in continual learning.
Across both benchmarks, conventional training rapidly leads to inactive neurons, while CBP and L2 init maintain balanced firing-rate distributions that closely resemble those of fully reset networks.

\section{Load balancing: controlling firing rates}
\label{sec:loadbalancing}

\Cref{sec:firing-rate} established two things: plasticity loss shows up as firing rates collapsing toward $0\%$ or saturating near $100\%$, and the marginal-entropy argument identifies an intermediate rate as the target.
Neither requires waiting for collapse to happen.
A firing rate is measured from the same tensor the activation already receives, and it can be moved by shifting that tensor --- so the target can be enforced continuously during training, at negligible cost and without touching the weights.

In this section, we investigate whether explicit control of firing rates can achieve performance beyond full network resets between tasks.
We introduce a lightweight mechanism, \emph{load balancing} (LB), that shifts each neuron's pre-activation to keep its long-term firing rate within target bounds.
Concretely, before a ReLU we replace the pre-activation $x$ by a shifted value
\begin{equation}
    \label{eq:shift}
    \tilde{x} = x + \sg(\beta),
\end{equation}
where ${\beta \in \R}$ is a per-neuron offset initialized to zero and $\sg(\cdot)$ denotes a stop-gradient (i.e., $\beta$ is treated as constant for backpropagation).
Rather than updating $\beta$ by gradient descent, we adjust it after each batch using a simple controller based on the exponentially averaged firing rate $\avgfiringrate$:
\begin{equation}
  \label{eq:update-rule}
  \beta \leftarrow \begin{cases}
    \beta + \alpha & \text{ if } \avgfiringrate < \minfiringrate \\
    \beta - \alpha & \text{ if } \avgfiringrate > \maxfiringrate \\
    \beta & \text{ otherwise,}
  \end{cases}
\end{equation}
where ${\alpha > 0}$ is a small step size (we use ${\alpha = 0.001}$) and ${[\minfiringrate, \maxfiringrate] \subset [0, 1]}$ specifies the target range.
Throughout, we write LB $a$--$b\%$ for the target range ${[a\%, b\%]}$, and LB $50\%$ for the single-rate target, which we implement as the narrow range 49--51\%.
We measure firing rates on the shifted pre-activations $\tilde{x}$ rather than on $x$, since $\tilde{x}$ is the signal the ReLU actually receives; the controller therefore closes the loop on the quantity it is regulating.
During inference, $\beta$ is fixed.
Intuitively, this shifts the effective activation threshold horizontally so that neurons spending too little (or too much) time in the active regime are nudged toward the target range.

\paragraph{Practical considerations.}
In practice, the shift $\beta$ can compete with the learned bias of the preceding layer.
We mitigate this by placing a BatchNorm immediately before the shift and using L2 regularization.
This configuration is already present in the class-incremental CIFAR-100 setup; for continual ImageNet we adopt the BN$\,+\,$L2 configuration.

\begin{algorithm}[b]
\caption{Load-balancing layer in PyTorch, for 1D inputs with $d$ features. \texttt{offset} is $\beta$, \texttt{avg\_rate} is $\avgfiringrate$.}
\label{alg:load-balancing}
\begin{mypython}[aboveskip=2pt, belowskip=0pt]
class LoadBalancer(nn.Module):
  def __init__(self, d, min_rate, max_rate, momentum=0.99, step=0.001):
    super().__init__()
    self.min_rate, self.max_rate = min_rate, max_rate
    self.momentum, self.step = momentum, step
    self.register_buffer("avg_rate", 0.5 * torch.ones(d))
    self.register_buffer("offset", torch.zeros(d))

  def forward(self, x):
    x = x + self.offset.unsqueeze(0)
    if self.training:
      with torch.no_grad():
        rate = (x > 0).float().mean(dim=0)
        self.avg_rate += (1 - self.momentum) * (rate - self.avg_rate)
        too_low = (self.avg_rate < self.min_rate).float()
        too_high = (self.avg_rate > self.max_rate).float()
        self.offset += self.step * (too_low - too_high)
    return x
\end{mypython}
\end{algorithm}

\paragraph{Implementation.}
We implement load balancing as a small layer inserted before each activation, which measures firing rates and adapts the shifts.
\Cref{alg:load-balancing} gives the layer for 1D inputs, e.g., after fully connected layers with $d$ features.

\needspace{6\baselineskip}
For 3D inputs, i.e., after convolutional layers, two lines of \cref{alg:load-balancing} change:
\begin{mypython}[belowskip=2pt]
x = x + self.offset.reshape(1, -1, 1, 1)
\end{mypython}
\begin{mypython}[aboveskip=2pt]
rate = (x > 0).float().mean(dim=[0, 2, 3])
\end{mypython}

\paragraph{Best practices.}
In our experiments, the EMA decay ${\avgmomentum = 0.99}$ and step size ${\alpha = 0.001}$ were robust across settings.
On the backbones of our main experiments, accuracy improves monotonically as the target range tightens (\cref{sec:training-details}), and the widest range we tried gains essentially nothing over the unconstrained reference.
We therefore recommend starting at a $50\%$ target (which we implement as 49--51\%) and loosening it only if training becomes unstable.
As noted above, we also recommend using load balancing together with a normalization layer placed before it.
We always insert the load-balancing layer immediately before the activation function; when a normalization layer is present, this means placing load balancing between the normalization and the activation.

\begin{figure}[t]
    \centerline{%
        \begin{minipage}[c]{0.5\textwidth}\centering%
            \includegraphics{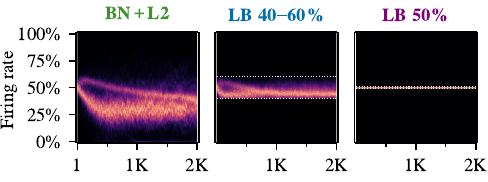}%
        \end{minipage}%
        \begin{minipage}[c]{0.5\textwidth}\centering%
            \includegraphics{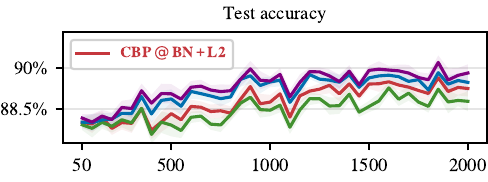}%
        \end{minipage}%
    }
    \caption{\textbf{Continual ImageNet:}
    Firing-rate distributions (left) and test accuracy (right) under BN$\,+\,$L2 with and without load balancing, averaged over 10 runs with $\pm 1$ standard error bands.
    Constraining neurons to intermediate regimes (40--60\% or 50\%) improves accuracy over both the unconstrained baseline and CBP.}
    \label{fig:imagenet-load-balancing}
\end{figure}
\begin{figure}[t]
    \centerline{%
        \begin{minipage}[c]{0.5\textwidth}\centering%
            \includegraphics{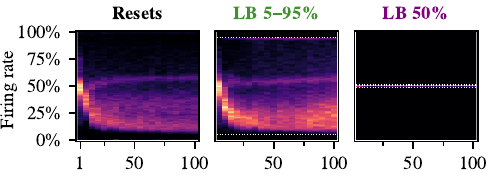}%
        \end{minipage}%
        \begin{minipage}[c]{0.5\textwidth}\centering%
            \includegraphics{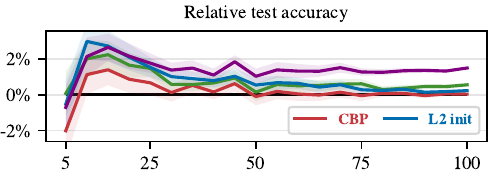}%
        \end{minipage}%
    }
    \caption{\textbf{Class-incremental CIFAR-100:}
    Firing-rate distributions (left) and test accuracy (right) for different load-balancing ranges, averaged over 15 runs with $\pm 1$ standard error bands; accuracy is relative to \emph{Resets}.
    Loose bounds (5--95\%) reproduce reset-like behavior with improved accuracy, while control near 50\% yields further gains, surpassing resets and prior methods.
    The LB 50\% panel concentrates nearly every neuron in a single bin, so it is dark apart from the line at 50\%.}
    \label{fig:cifar100-load-balancing}
\end{figure}

\paragraph{Load balancing in continual ImageNet.}
We use BN$\,+\,$L2 as the base configuration, since load balancing is more stable in the presence of normalization.
For comparability, we re-tune all baselines on this same setup; results are shown in \cref{fig:imagenet-more-baselines}, and hyperparameter details and full tables are provided in \cref{sec:training-details,sec:accuracy-tables}.
Throughout the paper we write ``method$\,\text{@}\,$reference'' for a method instantiated on a stated reference configuration, e.g.\ CBP$\,\text{@}\,$BN$\,+\,$L2 here and CBP$\,\text{@}\,$LN in \cref{sec:reinforcement-learning}.

On continual ImageNet, we evaluate load balancing under different firing rate bounds (\cref{fig:imagenet-load-balancing}).
Constraining firing rates to an intermediate band (40--60\%) yields higher accuracy than both unconstrained BN$\,+\,$L2 and CBP$\,\text{@}\,$BN$\,+\,$L2, and fixing all neurons to 50\% performs even better.

\paragraph{Load balancing in class-incremental CIFAR-100.}
In the class-incremental CIFAR-100 benchmark, we evaluate two load-balancing configurations and compare their performance relative to the Resets baseline, where the network is reinitialized after each task.
The results are presented in \cref{fig:cifar100-load-balancing}.
Using loose bounds of 5--95\% reproduces firing-rate patterns similar to the reset-based baseline.
This configuration already achieves higher accuracy than both resets and CBP.
Pushing the bounds more aggressively to 50\% yields even better performance.

\begin{figure}[t]
    \centerline{%
        \begin{minipage}[c]{0.5\textwidth}\centering%
            \includegraphics{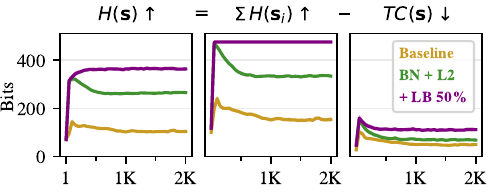}%
        \end{minipage}%
        \begin{minipage}[c]{0.5\textwidth}\centering%
            \includegraphics{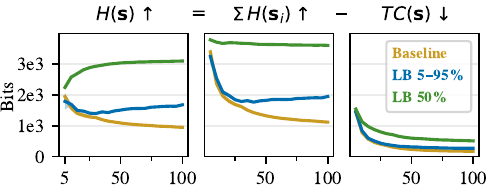}%
        \end{minipage}%
    }
    \caption{\textbf{Entropy:}
    Continual ImageNet (left) and class-incremental CIFAR-100 (right), with $H(\rvs)$, $\sum_i H(\rvs_i)$, and $TC(\rvs)$ computed on the test set, averaged over 3 random seeds.
    On continual ImageNet the three configurations are the Baseline, BN$\,+\,$L2, and BN$\,+\,$L2 with load balancing at $50\%$ (``$+$ LB 50\%''); on class-incremental CIFAR-100 they are the Baseline and load balancing at 5--95\% and 50\%.
}
    \label{fig:entropies}
\end{figure}

\paragraph{Measuring entropy.}
As discussed in \cref{sec:entropy-objective}, the joint entropy of the binary gates decomposes as
${H(\rvs) = \sum_{i=1}^M H(\rvs_i) - TC(\rvs)}$.
To monitor how the load-balancing mechanism affects these terms, we evaluate them on the test set after each task, using deterministic evaluation (e.g., BatchNorm in inference mode).
For each unit $i$, we estimate the marginal activation probability ${\Pr(\rvs_i = 1)}$ from the empirical firing rate on the test inputs, and compute the corresponding Bernoulli entropy $H(\rvs_i)$.
We report the sum $\sum_{i=1}^M H(\rvs_i)$ as a measure of the total marginal capacity; its maximum is $M$ bits, attained when every unit is active on exactly half of the inputs.
Note that these rates are measured at inference time, with frozen normalization statistics and a fixed $\beta$, so they need not coincide with the rates the controller maintains during training.
All entropy results are averaged over 3 random seeds.

Directly estimating the total correlation $TC(\rvs)$ is intractable for high-dimensional $\rvs$, as it would require the full joint distribution over $\{0,1\}^M$.
We therefore adopt a Chow--Liu tree approximation \citep{chow1968approximating}.
Concretely, we first estimate the pairwise mutual information $I(\rvs_i; \rvs_j)$ for all pairs $(i,j)$ from empirical joint activation counts.
We then construct a maximum spanning tree (MST), using $I(\rvs_i; \rvs_j)$ as edge weights.
For a distribution that factorizes according to this tree structure, the total correlation is
\begin{equation}
    TC(\rvs) \: = \!\!\!\! \sum_{(i,j) \:\in\: \mathrm{MST}} \!\!\!\! I(\rvs_i; \rvs_j),
\end{equation}
i.e., the sum of mutual informations along the tree edges.
We use this quantity as a tractable approximation that captures the dominant pairwise dependencies while ignoring higher-order interactions.

The resulting trajectories of $\sum_{i=1}^M H(\rvs_i)$ and the Chow--Liu estimate of $TC(\rvs)$ are shown in \cref{fig:entropies}.
Without load balancing, the sum of marginal entropies decays steadily over tasks.
Load balancing arrests that decay, and the tighter the target range the more of it is arrested: a $50\%$ target holds the sum near its initial level and close to its ceiling of $M$ bits, while 5--95\% falls part of the way.
The total-correlation term is higher under load balancing too, but by far less than the marginal entropies gain.
Because a tree captures only pairwise dependencies, the Chow--Liu estimate is a lower bound on the true $TC(\rvs)$, and ${\sum_{i=1}^M H(\rvs_i) - TC(\rvs)}$ is correspondingly an upper bound on the joint entropy $H(\rvs)$.
That bound is substantially higher with load balancing throughout training, consistent with the additional unit-wise capacity supporting more distinct affine regions rather than redundant neurons.
The three configurations also order the same way on entropy as on accuracy --- LB 50\%, then LB 5--95\%, then the unconstrained baseline (\cref{tab:cifar100-accuracy}) --- which is what the argument in \cref{sec:firing-rate} predicts.

\paragraph{Summary of findings.}

This section shows that explicitly controlling firing rates with load balancing improves continual-learning performance.
Constraining rates to intermediate ranges, e.g., 40--60\% or a fixed 50\%, yields higher accuracy than both the reset-based baseline and CBP on the two benchmarks.
These settings maintain diverse activation patterns and align with the earlier argument that a 50\% firing rate maximizes activation-state entropy and supports nonlinearity.
Consistent with this, our Chow--Liu estimates show that load balancing keeps the sum of marginal entropies far above the unconstrained baselines while raising redundancy by much less, so the upper bound on the joint entropy $H(\rvs)$ --- and with it the effective number of active regions --- stays correspondingly higher.
Taken together, the results highlight balanced neuron utilization as a useful target and indicate that firing-rate control is a practical mechanism for improving plasticity and knowledge transfer.

\section{Generality of load balancing}
\label{sec:generality}

So far we have applied load balancing to ReLU networks on two supervised continual-learning benchmarks.
We now ask how far the mechanism carries: to other activation functions, and to reinforcement learning.

\subsection{Beyond ReLU}
\label{sec:beyondrelu}

\begin{figure}
    \centerline{%
        \begin{minipage}[c]{0.5\textwidth}\centering%
            \includegraphics{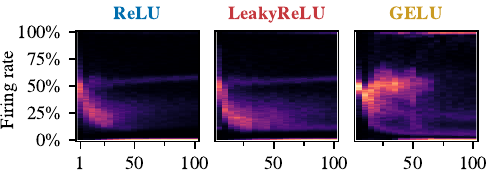}%
        \end{minipage}%
        \begin{minipage}[c]{0.5\textwidth}\centering%
            \includegraphics{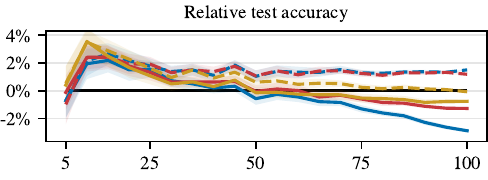}%
        \end{minipage}%
    }
    \caption{\textbf{Class-incremental CIFAR-100:}
    Firing-rate distributions (left) and test accuracy (right) for ReLU, LeakyReLU, and GELU, all based on the BN$\,+\,$L2 setup.  In the right plot the dashed lines show the results for load balancing 50\%.
}
    \label{fig:cifar100-activations}
\end{figure}

The main focus of this paper is on ReLU activations.
Nonetheless, it is also natural to ask how controlling the firing rate affects other activation functions.
We applied load balancing to LeakyReLU \citep{maas2013rectifier} and GELU \citep{hendrycks2016gaussian}.
The results are shown in \cref{fig:cifar100-activations,fig:imagenet-activations}.
In each figure, the left panels show the firing-rate distributions without load balancing, while the right panels compare performance without load balancing (solid lines) and with load balancing at 50\% (dashed lines).

On class-incremental CIFAR-100 (\cref{fig:cifar100-activations}), all three activation functions lose plasticity, and all three benefit from load balancing at 50\%.
The gain is largest for ReLU: without load balancing it is the weakest of the three, and with it the strongest, on par with load-balanced LeakyReLU.

On continual ImageNet (\cref{fig:imagenet-activations}), none of the three shows substantial plasticity loss to begin with, because that setup already uses BatchNorm and L2 regularization.
With each activation at its own tuned learning rate and L2 strength, all three still benefit on all 10 task sequences: comparing each against its own unbalanced counterpart gives ${+0.99 \pm 0.05}$ for ReLU, ${+0.88 \pm 0.11}$ for LeakyReLU and ${+0.31 \pm 0.05}$ for GELU.
GELU, the strongest without load balancing, gains least; with it, all three end within $0.12$ points.

\subsection{Reinforcement learning}
\label{sec:reinforcement-learning}

Beyond supervised continual learning, we also tested our load-balancing mechanism in a reinforcement-learning setting.
We conducted experiments with Rainbow DQN \citep{hessel2018rainbow} on the Atari-5 benchmark \citep{aitchison2022atari5} and on \textit{DemonAttack}, an environment previously used to demonstrate plasticity loss \citep{sokar2023dormant}.
We consider four agents: (i) the original Rainbow DQN baseline, (ii) Rainbow DQN with LayerNorm in the hidden layers, (iii) the same architecture further augmented with our load-balancing layers targeting a 50\% firing rate (LB 50\%$\,\text{@}\,$LN), and (iv) LayerNorm with CBP (CBP$\,\text{@}\,$LN).
We found that disabling load balancing for the value head consistently improved performance, so all reported results use this configuration.
Other than these architectural changes, we keep the Rainbow hyperparameters fixed and use the Adam optimizer; full details are given in \cref{sec:training-details}.
Performance is reported in terms of \emph{human-normalized scores} (HNS), computed as
\begin{equation}
    \mathrm{HNS} = \frac{\mathrm{score}_{\text{agent}} - \mathrm{score}_{\text{random}}}{\mathrm{score}_{\text{human}} - \mathrm{score}_{\text{random}}},
\end{equation}
so that $0$ corresponds to a random agent and $1$ to a human expert.

\begin{figure}[t]
    \centerline{\includegraphics{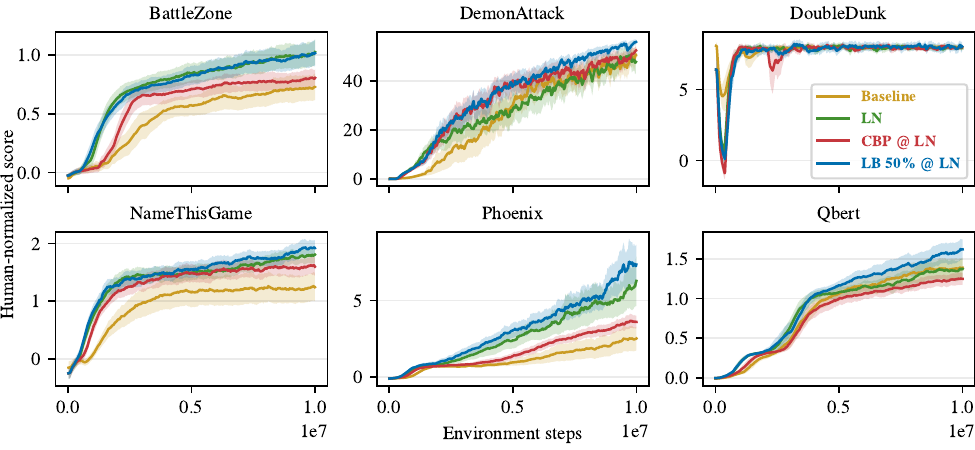}}
    \caption{\textbf{Rainbow DQN:}
    Human-normalized scores (0 = random, 1 = human) on 6 Atari games for four variants: the original baseline, a baseline with LayerNorm (LN), and load balancing at a 50\% target and CBP instantiated on that LayerNorm agent (LB 50\%$\,\text{@}\,$LN and CBP$\,\text{@}\,$LN).
    Curves show the mean over 5 random seeds and are smoothed with a time-weighted exponential moving average.}
    \label{fig:rainbow-results}
\end{figure}

\begin{figure}[t]
    \centerline{\includegraphics{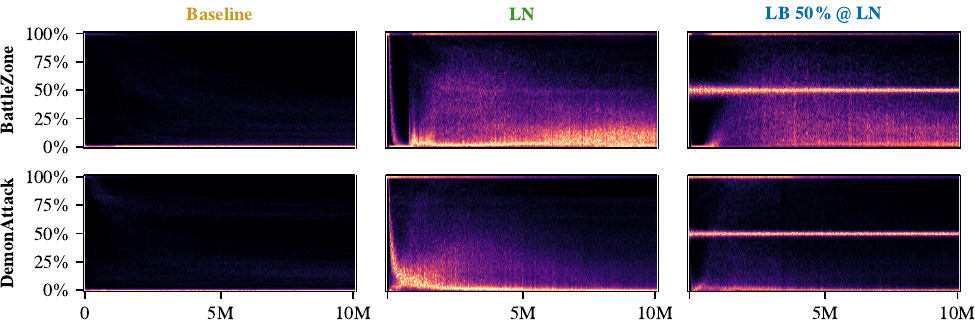}}
    \caption{\textbf{Rainbow DQN:} Firing-rate distributions, one row per game and one column per agent, in the same order as \cref{fig:rainbow-results}.
    The baseline collapses to very low rates in both games; LN lifts them part of the way; load balancing holds the hidden layers near the $50\%$ target, with the residual mass at $0\%$ and $100\%$ coming from the value head, which is excluded by design.
    The contrast between the two games explains the accuracy result: LayerNorm alone already raises firing in \textit{BattleZone}, leaving load balancing little to add, while in \textit{DemonAttack} it does not.}
    \label{fig:rainbow-rates}
\end{figure}

Across the 6 games, the load-balancing agent attains the highest human-normalized score of the four agents in 4 games and ties for the highest in the remaining 2 (\cref{fig:rainbow-results,tab:rainbow-scores}).
The original Rainbow baseline remains competitive in some games --- it is on par with all other agents in \textit{DoubleDunk} and with LayerNorm in \textit{DemonAttack} and \textit{Qbert} --- but is worse than load balancing in 5 of 6 games.
Load balancing outperforms CBP in 5 of 6 games and matches it in \textit{DoubleDunk}.

The firing-rate distributions in \cref{fig:rainbow-rates} mirror these performance trends.
In the original Rainbow baseline, almost all neurons fire at very low rates, indicating severe collapse.
Adding LayerNorm partially alleviates this, but most firing rates still remain far below 50\%.
With load balancing, by contrast, most neurons in the hidden layers operate close to the 50\% target (the value head is excluded by design).
The distributions also clarify why load balancing helps in \textit{DemonAttack} but not in \textit{BattleZone}: in \textit{BattleZone}, the LayerNorm baseline already causes more neurons to fire more frequently, leaving less room for improvement, whereas in \textit{DemonAttack} the LayerNorm baseline remains highly collapsed and load balancing produces a much larger shift in firing rates.

\section{Firing rates as a diagnostic}
\label{sec:diagnostic}

The preceding sections used firing rates to motivate and evaluate a mechanism.
This section turns the lens around and asks what the diagnostic itself is worth: where in a network collapse actually happens, how firing rates compare with the plasticity metrics already in the literature, and whether they are informative enough to guide architecture design.

\subsection{Layer-wise firing rates}
\label{sec:layer-wise-analysis}

Thus far we have visualized firing rates aggregated over all neurons.
Layer-wise views reveal where collapse or linearization emerges.
All heatmaps follow the same visualization convention as before.

\begin{figure}[t]
    \centerline{\includegraphics{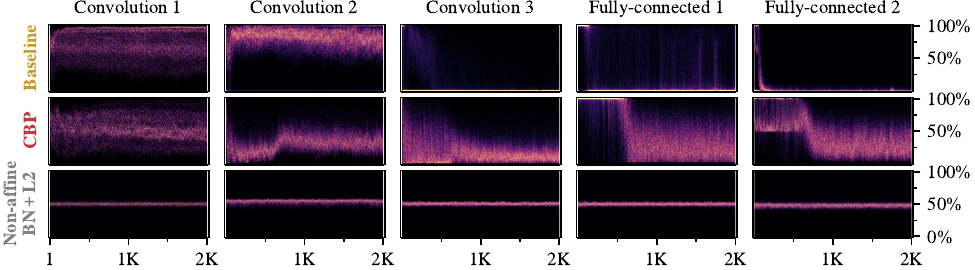}}
    \caption{\textbf{Continual ImageNet:}
    Layer-wise firing-rate distributions over tasks.
    Rows correspond to methods, columns correspond to successive layers.
    Early layers stay active while later layers collapse in the baseline; CBP shows heterogeneous behavior; non-affine BN$\,+\,$L2 maintains near-balanced firing rates across layers.}
    \label{fig:imagenet-layerwise-rates}
\end{figure}

\begin{figure}[t]
    \centerline{\includegraphics{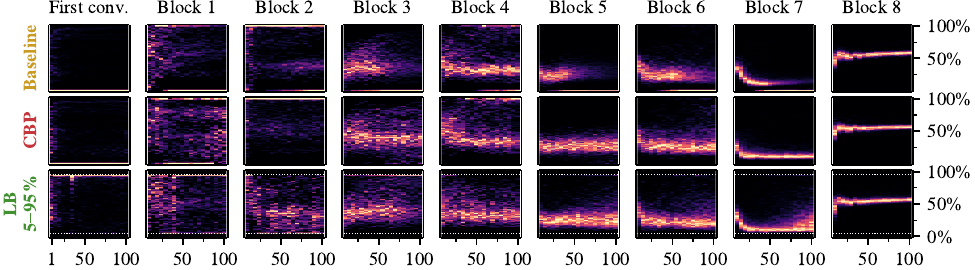}}
    \caption{\textbf{Class-incremental CIFAR-100:}
    Layer-wise firing-rate distributions over tasks.
    Continual training and CBP exhibit linearization in Block 2.
    Load balancing with a $95\%$ upper limit prevents it, at the cost of some saturation at the limit in the early layers.}
    \label{fig:cifar100-layerwise-rates}
\end{figure}

In \cref{fig:imagenet-layerwise-rates}, we show layer-wise firing-rates dynamics for continual ImageNet.
In the first row (baseline), the first two convolutional layers sustain high activity while subsequent layers collapse toward $0\%$.
In the second row (CBP), activity varies considerably across layers over time.
By contrast, non-affine BN$\,+\,$L2 (third row) maintains stable, near-balanced firing rates across layers.
Comprehensive results for all approaches are provided in \cref{fig:imagenet-all-layerwise-rates}.

In \cref{fig:cifar100-layerwise-rates} (class-incremental CIFAR-100), we observe linearization emerging in the second ResNet block (Block 2) under continual training and CBP.
Load balancing with an upper limit of $95\%$ prevents this effect, which may contribute to its higher test accuracy.
However, under load balancing, firing rates sometimes accumulate at the $95\%$ limit in early layers, indicating saturation that is also suboptimal.
Comprehensive results for all approaches are provided in \cref{fig:cifar100-all-layerwise-rates}.

\subsection{Relation to existing plasticity metrics}
\label{sec:comparison}

Firing rates are not the only diagnostic that has been proposed for plasticity loss.
This section relates them to three quantities from the literature -- the fraction of dormant neurons, weight magnitudes, and the spectral norm together with the effective rank -- measured on the same runs as the results above.
As with the accuracy results, each curve is a mean over $5$ runs with $\pm 1$ standard error bands, so the orderings we read off below can be checked against the seed-to-seed spread.

\begin{figure}[t]
    \centerline{\includegraphics{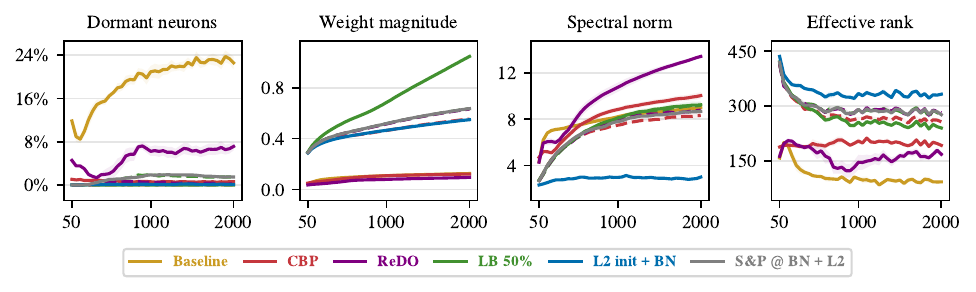}}
    \caption{\textbf{Continual ImageNet:}
    From left to right: fraction of dormant neurons (threshold ${\tau_\text{d} = 0.01}$), weight magnitudes, overall maximum spectral norm across layers, and effective rank \citep{lewandowski2025learning}, over tasks.
    Dashed lines indicate variants built on a BatchNorm backbone (BN$\,+\,$L2, except L2 init$\,+\,$BN, which adds no further L2).
    Averaged over $5$ runs with $\pm 1$ standard error bands.}
    \label{fig:imagenet-metrics}
\end{figure}

\begin{figure}[t]
    \centerline{\includegraphics{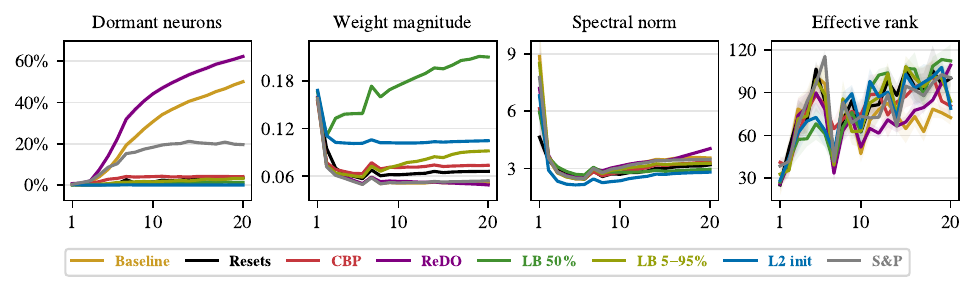}}
    \caption{\textbf{Class-incremental CIFAR-100:}
    From left to right: fraction of dormant neurons (threshold ${\tau_\text{d} = 0.01}$), weight magnitudes, overall maximum spectral norm across layers, and effective rank \citep{lewandowski2025learning}, over tasks.
    Averaged over $5$ runs with $\pm 1$ standard error bands.}
    \label{fig:cifar100-metrics}
\end{figure}

\paragraph{Dormant neurons.}
A common proxy for plasticity is the fraction of \emph{dormant} neurons \citep{sokar2023dormant}.
Following their definition, a neuron's \emph{score} is its activation magnitude, normalized by the magnitudes of other neurons in the same layer and averaged over the dataset; a neuron is labeled dormant if its score falls below a threshold $\tau_\text{d}$ (we use ${\tau_\text{d} = 0.01}$).
In \cref{fig:imagenet-metrics,fig:cifar100-metrics}, most approaches maintain a near-zero fraction of dormant units relative to their respective baselines.
Two approaches are exceptions, both on class-incremental CIFAR-100 and both by margins far larger than the seed spread.
ReDO ends at ${62.3 \pm 0.7\%}$ dormant units, above the continually trained baseline at ${50.0 \pm 0.9\%}$.
S\&P ends at ${19.5 \pm 0.5\%}$, well below the baseline but nearly five times the next largest, as no remaining approach exceeds ${5\%}$.
Its firing-rate distribution already indicated this, since in \cref{fig:cifar100-firing-rates} S\&P shifts the distribution toward the \emph{Resets} regime while many neurons remain near ${\approx 0\%}$.
Overall, $0\%$ firing is a sufficient (though not necessary) condition for dormancy, explaining the observed alignment between the metrics.
Dormancy primarily captures the low-activity neurons.
Firing-rate distributions are broader, as they reveal low-rate collapse, high-rate linearization, and intermediate-range activity.

\paragraph{Weight magnitudes.}
Plasticity loss is often reported to be accompanied by growth in weight norms \citep{dohare2024loss}.
We do not observe that association.
On continual ImageNet the weight magnitudes separate the two backbones rather than the approaches.
The three approaches without BatchNorm end below ${0.13}$, all six built on it above ${0.54}$, and neither group has a consistent internal ordering.
The gap is not an L2 effect, since the approaches that end lowest are precisely those trained without weight decay.
On class-incremental CIFAR-100, where every approach shares one architecture and one weight decay, the association is reversed.
The two most dormant approaches, ReDO and the baseline, end with the smallest weight magnitudes, and the two least dormant, L2 init and LB 50\%, with the largest (Spearman ${\rho = -0.95}$, ${p < 0.001}$, over all eight approaches).
Dormant neurons receive no gradient, so weight decay shrinks their weights unopposed, and at fixed architecture and fixed L2 the weight magnitude reflects how much of the network is still active.
This agrees with \citet{lyle2024disentangling}, who find no monotone relationship between the parameter norm at the end of training and plasticity, and note that networks whose units are saturated do not propagate the gradients that would contribute to norm growth.
Load balancing fits this trend rather than breaking it, and the bias growth it induces in the preceding BatchNorm layers (the shift competes with the learned bias) puts LB 50\% furthest out.

\paragraph{Spectral norm and effective rank.}
\citet{lewandowski2025learning} have shown that the spectral norm of the network layers and the effective rank are related to plasticity.
In \cref{fig:imagenet-metrics,fig:cifar100-metrics}, we evaluate the overall maximum spectral norm across layers, and the effective rank.
On continual ImageNet the effective rank does order the approaches.
The baseline ends with by far the lowest value (${92 \pm 5}$), while the variants built on BN$\,+\,$L2 all retain between ${257}$ and ${333}$, a gap the seeds resolve comfortably.
The spectral norm is less consistent.
Its largest value belongs to ReDO, which is also the second most dormant approach, but the next largest belong to approaches with essentially no dormant units, and the baseline -- the most dormant of all -- ranks only fifth, within one standard error of the three approaches that bracket it.
On class-incremental CIFAR-100 the two metrics come apart.
The effective rank does not order the approaches at all, since the curves cross repeatedly from task to task and six of the seven gaps between consecutively ranked methods are smaller than the corresponding standard errors.
The spectral norm does produce an ordering here, and one that tracks dormancy almost exactly, with ReDO and the baseline highest and the approaches that keep neurons active lowest.
That ordering is the reverse of what continual ImageNet showed, and it is compressed into the range ${2.8}$ to ${4.0}$ against ${3.0}$ to ${13.5}$ there, so it rests on separations the seed bands resolve rather than on the visible spread between curves.
Neither of these two metrics behaves the same way on both benchmarks either, whereas firing-rate distributions remain informative in both settings.
This is what we want from a diagnostic: it should not stop discriminating, or change what it discriminates on, when the benchmark changes.

\subsection{Case study: designing balanced networks}
\label{sec:case-study}

The preceding sections showed that plasticity correlates with non-collapsed firing-rate distributions, with the best results near 50\%.
While load balancing enforced this explicitly, we now ask whether architecture alone can induce the same behavior.
Guided by firing-rate diagnostics, tracking layer-wise distributions, we identify minimal, effective modifications.
The goal is not exhaustive architecture design, but a targeted case study of how the firing-rate lens translates into practical design choices.

\begin{figure}[t]
    \centerline{%
        \begin{minipage}[c]{0.5\textwidth}\centering%
            \includegraphics{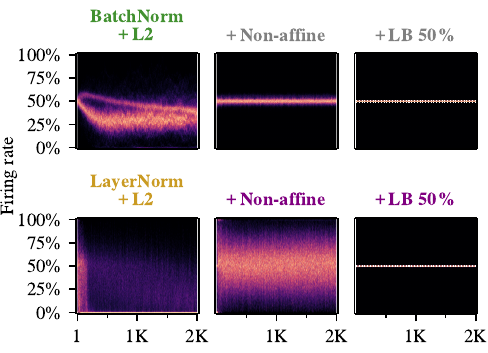}%
        \end{minipage}%
        \begin{minipage}[c]{0.5\textwidth}\centering%
            \includegraphics{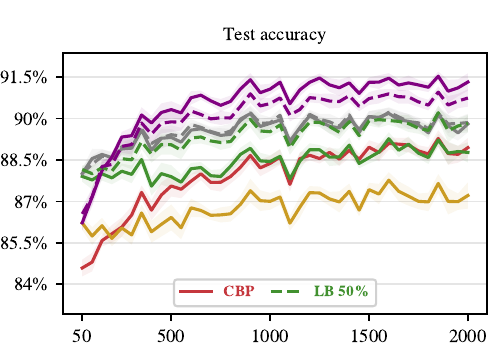}%
        \end{minipage}%
    }
    \caption{\textbf{Continual ImageNet:}
    Effects of non-affine normalization on firing-rate distributions (left) and test accuracy (right).
    Non-affine BatchNorm$\,+\,$L2 clusters firing rate near 50\%, while non-affine LayerNorm$\,+\,$L2 broadens the distribution across layers.
    Both configurations improve accuracy, with non-affine LayerNorm achieving the highest performance of all methods.
    Enforcing $50\%$ via load balancing on top of non-affine LayerNorm tightens rates but lowers accuracy.
    }
    \label{fig:imagenet-techniques}
\end{figure}

\begin{figure}
    \centerline{%
        \begin{minipage}[c]{0.5\textwidth}\centering%
            \includegraphics{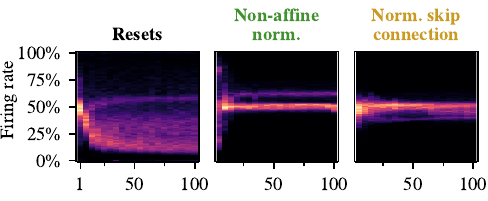}%
        \end{minipage}%
        \begin{minipage}[c]{0.5\textwidth}\centering%
            \includegraphics{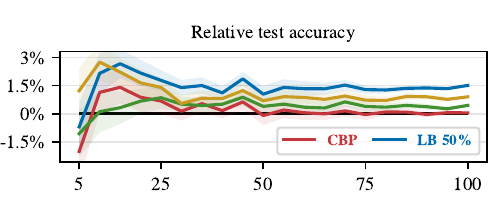}%
        \end{minipage}%
    }
    \caption{\textbf{Class-incremental CIFAR-100:}
    Firing-rate distributions (left) and test accuracy (right) under non-affine normalization and normalized skip connections; accuracy is relative to \emph{Resets}, as in \cref{fig:cifar100-load-balancing}.
    Both modifications steer firing rates toward balanced regimes and improve accuracy beyond resets and CBP.
}
    \label{fig:cifar100-techniques}
\end{figure}

\paragraph{Non-affine normalization.}
A direct way to keep ReLU units near the 50\% firing regime is to keep their pre-activations centered at the threshold (zero).
Normalization layers compute a standardized signal and then apply a learned affine transform with scale $\gamma$ and shift $\delta$:
\begin{equation}
    \hat{x} = \frac{x - \mathbb{E}[x]}{\sqrt{\text{Var}(x) + \epsilon}} \cdot \gamma + \delta.
\end{equation}
BatchNorm \citep{ioffe2015batch} estimates the mean and variance over the batch; LayerNorm \citep{ba2016layer} does so over features.
Crucially, the learned shift $\delta$ can move the normalized signal away from zero, so these layers generally do not keep the activation centered at the ReLU threshold.

Therefore, we consider non-affine variants that drop the learned scale $\gamma$ and shift $\delta$.
As shown in \cref{fig:imagenet-techniques}, non-affine BatchNorm$\,+\,$L2 yields firing rates tightly clustered around $\approx 50\%$ and outperforms CBP$\,\text{@}\,$BN$\,+\,$L2 on all 10 task sequences (${+0.38 \pm 0.10}$).
In contrast, non-affine LayerNorm$\,+\,$L2 produces firing rates broadly distributed across 0--100\% and achieves the highest accuracy on this benchmark, surpassing even our load balancing results.
This indicates that while centering promotes balanced activity, performance gains can also arise from wider, layer-wise variability in firing rates.

Adding load balancing that targets a strict $50\%$ firing rate on top of non-affine LayerNorm reduces accuracy (${-0.53 \pm 0.06}$, see \cref{fig:imagenet-techniques}), whereas the same target on non-affine BatchNorm slightly improves it (${+0.19 \pm 0.06}$): the sign depends on the normalization, not on the target.
This suggests that enforcing a single rate across all units is not universally optimal; where balanced utilization is already distributed across the full range of firing rates, varying by layer and unit, tightening it costs accuracy.
In \cref{fig:cifar100-techniques}, non-affine BatchNorm also improves accuracy on class-incremental CIFAR-100 and already exceeds the reset-based baseline and CBP. The corresponding firing-rate distributions are concentrated in an intermediate range rather than collapsing to 0\% or saturating near 100\%.

\paragraph{Normalized skip connections.}
Building on the normalization view, \cref{fig:cifar100-skip-normalize} plots the firing rates of the first and second ReLU in each ResNet block for class-incremental CIFAR-100.
Despite using non-affine BatchNorm, the distribution of the second ReLU is skewed (second panel).
The cause is architectural: following the classical post-activation block of \citet{he2016deep} as used by \citet{dohare2024loss} (\texttt{Conv-BatchNorm-ReLU-Conv-BatchNorm}, skip connection, second ReLU), the pre-activation to the second ReLU is \emph{not} normalized.
In the ``Norm. skip connection'' variant, we apply a non-affine BatchNorm to the skip path.
The right panels of \cref{fig:cifar100-skip-normalize} show that this yields a more symmetric distribution around $50\%$.
\cref{fig:cifar100-techniques} confirms the same trend network-wide, together with a gain in accuracy.
While it is known that the original post-activation block has drawbacks, our results provide a complementary explanation through the lens of firing-rate balance.

\paragraph{Summary of findings.}
Non-affine normalization layers and normalized skip connections shift firing-rate distributions toward balanced regimes, often near 50\%, and consistently improve accuracy in continual benchmarks. While their benefits have been discussed from other perspectives, viewing them through firing-rate dynamics highlights why such modifications improve performance.

\section{Limitations}
\label{sec:limitations}

For the ReLU activations (and to some degree also for LeakyReLU and GELU, see \cref{sec:beyondrelu}) the firing rate is easily definable via the sign of the pre-activation.
However, for saturating symmetric activations such as tanh or sigmoid it is less clear how the rate should be controlled.
Options include controlling the amount of saturation (large absolute pre-activations) and the degree of linearity (pre-activations around zero).
We leave it to future work to link the entropy-based view to these cases.

Our load-balancing layer is designed to operate in conjunction with normalization layers, which stabilize pre-activations so that firing-rate shifts are meaningful.
Methods like CBP are more agnostic to the underlying architecture.
Nonetheless, load balancing is simple to implement, adds only one non-trainable offset per neuron and an elementwise shift, and integrates naturally into widely used architectures such as ResNets.

Our evidence is also bounded in ways worth naming.
The continual ImageNet effects are small in absolute terms --- around one accuracy point --- and are resolved only because every method is evaluated on the same task sequences; at this sample size we would not expect them to be visible in an unpaired comparison.
The plasticity metrics in \cref{sec:comparison} are averaged over $5$ runs rather than the $10$ and $15$ used for accuracy, so the smallest of the orderings we read off them are supported by correspondingly wider error bands.
The entropy analysis relies on a Chow--Liu approximation that bounds the joint entropy from above rather than estimating it, so it shows that the additional unit-wise capacity is not offset by redundancy, not that the joint entropy attains any particular value.
Finally, a fixed $50\%$ target is not universally best: \cref{sec:case-study} finds a configuration whose firing rates are spread across the full range and which outperforms it, so the marginal-entropy argument is a useful guide rather than a complete account of what balanced utilization buys.

\section{Conclusion}

This work identifies neuron firing rates as a simple, interpretable signal for diagnosing and improving plasticity in continual learning.
We show that continual training often drives rates to extremes, producing dead or effectively linear units.
To counter this, we introduce a lightweight load-balancing mechanism that maintains firing rates within a target range and outperforms the reset-based methods on both supervised benchmarks and on Atari.
We further demonstrate that architectural choices can implicitly encourage a similar balance --- on continual ImageNet, well enough to beat explicit control.
Taken together, these results position firing-rate distributions as a practical design tool for building more robust continual learners.

\section*{Acknowledgments}

This research has been funded and supported by the Federal Ministry of Research, Technology and Space of Germany and the state of North Rhine-Westphalia as part of the Lamarr Institute for Machine Learning and Artificial Intelligence.

\section*{Broader impact statement}

This work studies why neural networks lose the ability to keep learning as their data distribution shifts, and proposes a lightweight mechanism that mitigates it.
Sustaining plasticity reduces the need to retrain models from scratch whenever data changes, which lowers the compute and energy cost of keeping deployed models current; it also makes it easier for a model to keep absorbing new data, so the usual considerations around what a continually updated model is trained on continue to apply.
We see no consequences specific to this work that call for further discussion.

\section*{Reproducibility statement}

Our experiments build on the official implementation of \citet{dohare2024loss}\footnote{\url{https://github.com/shibhansh/loss-of-plasticity}}.
On class-incremental CIFAR-100 we use the hyperparameters reported in their paper.
On continual ImageNet we retune the learning rate and the regularization strengths for each method, since our configurations differ from theirs; all sweep ranges and selected values are given in \cref{sec:training-details}.
Beyond this we make two deviations. First, we add our load-balancing layer before each ReLU activation (implementation details are in \cref{sec:loadbalancing}). Second, when using L2 regularization we apply it to weights only, excluding biases and normalization-layer parameters (see \cref{sec:training-details} for more details).
All new hyperparameters are specified in the main text.

\section*{Use of large language models (LLMs)}

Large language models (LLMs) were used as assistive tools in preparing this paper.
For the writing, they helped with phrasing, clarity and figure captions, and proposed revisions of individual paragraphs.
They were also consulted for exposition of background concepts such as entropy and mutual information.
For the analysis, LLM coding assistants helped write the scripts that produce the tables and figures from the logged training runs, and check the reported numbers against those runs.
They also helped plan and specify the additional hyperparameter tuning runs.
The research questions, the load-balancing method and the theoretical arguments are the authors' own.
The authors reviewed all LLM-generated text and code and take full responsibility for the content of this paper.

\bibliography{main}
\bibliographystyle{tmlr}

\newpage
\appendix

\section{Additional results}

\begin{figure}[h!]
    \centerline{\includegraphics{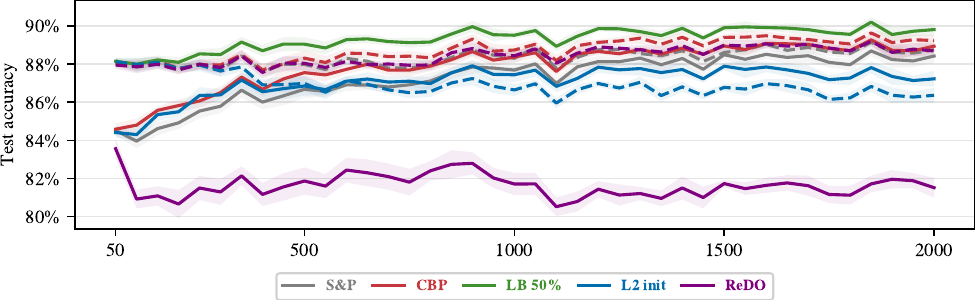}}
    \caption{\textbf{Continual ImageNet:}
    Test accuracy of additional baselines.
    Dashed lines indicate variants built on a BatchNorm backbone (BN$\,+\,$L2, except L2 init$\,+\,$BN, which adds no further L2).
    }
    \label{fig:imagenet-more-baselines}
\end{figure}

\begin{figure}[h!]
    \centerline{%
        \begin{minipage}[c]{0.5\textwidth}\centering%
            \includegraphics{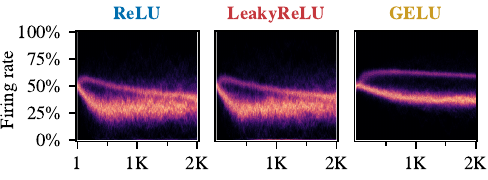}%
        \end{minipage}%
        \begin{minipage}[c]{0.5\textwidth}\centering%
            \includegraphics{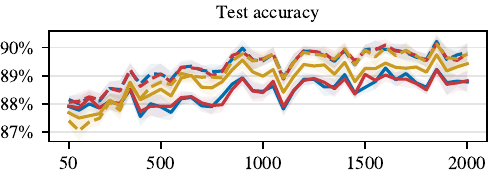}%
        \end{minipage}%
    }
    \caption{\textbf{Continual ImageNet:}
    Firing-rate distributions (left) and test accuracy (right) for ReLU, LeakyReLU, and GELU, all based on the BN$\,+\,$L2 setup.  In the right plot the dashed lines show the results for load balancing at 50\%.}
    \label{fig:imagenet-activations}
\end{figure}

\begin{figure}[h!]
    \centerline{\includegraphics{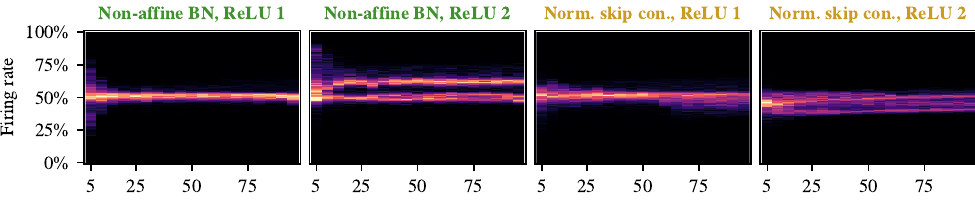}}
    \caption{\textbf{Class-incremental CIFAR-100:}
    Firing-rate distributions of the first and second ReLU within all ResNet blocks.
    \textbf{Left:} Non-affine BatchNorm baseline; the second ReLU distribution is skewed.
    \textbf{Right:} Non-affine BatchNorm on the skip path before addition re-centers the second ReLU pre-activation, yielding a more symmetric distribution around 50\%.}
    \label{fig:cifar100-skip-normalize}
\end{figure}

\section{Extended theoretical discussion}
\label{sec:further-theory}

\paragraph{Beyond continuous piecewise-affine networks.}
When a network is built from affine layers together with piecewise-linear operations (e.g., ReLU and max-pooling), the overall input-output map is continuous piecewise affine (CPWA), so activation patterns index affine regimes of the computation \citep{montufar2014linear,serra2018bounding,goujon2024linear}.
This includes common CNNs and ResNets with BatchNorm at inference time: with frozen running statistics, BatchNorm reduces to an affine transformation and therefore preserves piecewise linearity structure \citep{bunel2018unified,balestriero2022batch}.
In architectures employing smooth normalizations such as LayerNorm, the map is no longer piecewise affine, since LayerNorm normalizes by a per-example standard deviation (involving a square root and division), introducing a smooth nonlinearity even if the surrounding layers are linear \citep{ba2016layer,ni2024nonlinearity}.
In that case, $S$ should be interpreted as indexing \emph{operating regimes} or effective subnetworks (which ReLU branches are taken, and hence which local computation is realized) rather than exact affine regions.
Nonetheless, the coding viewpoint remains useful: entropy or mutual-information objectives over activation codes are motivated as a way to encourage diverse, noise-robust internal regimes independent of strict CPWA assumptions \citep{park2021unsupervised}.

\paragraph{Average Hamming distance.}
Here we make explicit how our marginal-entropy objective relates to the average Hamming distance between binary activation codes.
Let ${p_i \coloneqq \Pr(\rvs_i = 1)}$, and consider two independent inputs ${x, x' \sim P_\rvx}$ with codes $S(x)$ and $S(x')$.
Their expected Hamming distance decomposes as
\begin{align}
    \label{eq:avg-hamming}
    \bar d_H
    & \coloneqq \E_{x,x'}\bigl[d_H\bigl(S(x), S(x')\bigr)\bigr] \\
    & = \sum_{i=1}^M \Pr\bigl(\rvs_i(x) \neq \rvs_i(x')\bigr) \\
    & = \sum_{i=1}^M 2 \, p_i (1 - p_i).
\end{align}
Both the Bernoulli entropy $H(\rvs_i)$ and the variance term ${p_i (1-p_i)}$ are maximized at ${p_i = 1/2}$, and are strictly decreasing as $p_i$ moves towards $0$ or $1$.
Thus, maximizing ${\sum_i H(\rvs_i)}$ not only avoids dead or saturated units, but also increases the average Hamming distance $\bar d_H$ between activation patterns, making codes for different inputs more widely separated.
This perspective is closely related to the noisy-channel analysis of \citet{park2021unsupervised}, who show that in their Neural Activation Coding objective the mutual information between inputs and noisy binary codes lower-bounds the average Hamming distance between codewords, so maximizing mutual information yields maximally separated, noise-robust activation codes.

\section{Training details}
\label{sec:training-details}

\revision{
\paragraph{Continual ImageNet.}
The model is a compact CNN consisting of three convolutional layers followed by two fully connected layers with ReLU activations.
L2 regularization is applied to weights only, excluding biases and parameters of normalization layers.
This differs from the original CBP codebase but follows common practice and was necessary for strong L2 performance.
All reported accuracies are averaged over 10 runs, which differ in the sampled task sequence; every method is evaluated on the same 10 sequences.
The plasticity metrics of \cref{sec:comparison} require additional per-task logging and are averaged over 5 runs, drawn the same way.
All experiments were run on a single NVIDIA A100, H100 or H200 GPU with \texttt{torch.compile} and, except for CBP, fp16 mixed precision. A full run takes roughly $2.8$ hours on an A100 for most methods; measured wall-clock medians are reported at the end of this section.

All hyperparameters and sweep ranges are reported in \cref{tab:imagenet-hyperparameters}; set-valued entries indicate the values considered in our sweeps.
We tune two \emph{reference configurations} that serve as starting points for subsequent experiments: \textbf{Baseline} (no BatchNorm, no L2) and \textbf{BN + L2} (both enabled).
For normalization- and regularization-based additions (\emph{BN}, \emph{L2}, \emph{S\&P}, \emph{L2 init}, \emph{non-affine normalization}, and their combinations), we jointly sweep the learning rate and the relevant regularization hyperparameters (L2 strength where applicable, the noise scale for S\&P, and the regularization strength for L2 init), selecting the setting that maximizes accuracy averaged over the final 100 tasks.
We then re-run the selected setting with the full number of runs reported in the tables, and report those runs.
The best-performing hyperparameters are:
\begin{itemize}[topsep=0pt,itemsep=0pt]
    \item \textbf{Baseline:} learning rate $0.001$
    \item \textbf{BN only:} learning rate $0.01$
    \item \textbf{L2 only:} learning rate $0.03$, L2 strength $3\mathrm{e}{-06}$
    \item \textbf{S\&P:} learning rate 0.01, L2 strength $1\mathrm{e}{-05}$, noise scale $1\mathrm{e}{-05}$
    \item \textbf{S\&P @ BN + L2:} learning rate 0.03, L2 strength $1\mathrm{e}{-06}$, noise scale $1\mathrm{e}{-06}$
    \item \textbf{L2 init:} learning rate 0.01, L2 init strength $1\mathrm{e}{-05}$
    \item \textbf{L2 init + BN:} learning rate 0.03, L2 init strength $1\mathrm{e}{-05}$
    \item \textbf{BN + L2:} learning rate $0.03$, L2 strength $1\mathrm{e}{-06}$
    \item \textbf{Non-affine BN + L2:} learning rate $0.1$, L2 strength $1\mathrm{e}{-06}$
    \item \textbf{Non-affine LN + L2:} learning rate $0.1$, L2 strength $1\mathrm{e}{-06}$
\end{itemize}
The two naming patterns in this list are deliberate: S\&P$\,\text{@}\,$BN$\,+\,$L2 is instantiated on the tuned BN$\,+\,$L2 reference and keeps its L2 strength, whereas L2 init$\,+\,$BN combines L2 init with BatchNorm and uses no additional L2 --- the regularization toward the initialization takes its place.
For methods that introduce their own hyperparameters (\emph{ReDO}, \emph{CBP}, \emph{Load balancing}), we instantiate them on top of a tuned reference configuration and keep the reference hyperparameters fixed (learning rate and, if present, L2 strength).
We then sweep only the method-specific hyperparameters and again select by accuracy over the last 100 tasks.

The best-performing method-specific hyperparameters are:
\begin{itemize}[topsep=0pt,itemsep=0pt]
    \item \textbf{ReDO @ Baseline:} dormancy threshold 0.1, recycling period 10
    \item \textbf{ReDO @ BN + L2:} dormancy threshold 0, recycling period 10
    \item \textbf{CBP @ Baseline:} replacement rate 0.0003
    \item \textbf{CBP @ BN + L2:} replacement rate 0.0003
    \item \textbf{Load balancing @ BN + L2:} target range 50\%
\end{itemize}
GELU and LeakyReLU are tuned like the other variants, jointly over learning rate and L2 strength on the BN$\,+\,$L2 backbone (best: learning rate $0.01$ and $0.03$ respectively, both with L2 strength $1\mathrm{e}{-06}$), and load balancing is added at those settings.

\paragraph{Class-incremental CIFAR-100.}

Each task begins with five randomly sampled classes from CIFAR-100.
Every 200 epochs, five additional classes are introduced, and the model is trained to classify all classes encountered so far.
The architecture is a ResNet-18 with ReLU activations, adapted to $32 \times 32$ CIFAR-100 images.
Each new task starts from the best checkpoint of the previous task, selected using validation accuracy, effectively applying early stopping \citep{dohare2024loss}.
All reported accuracies are averaged over 15 random seeds.
The plasticity metrics of \cref{sec:comparison} require additional per-task logging and are averaged over 5 seeds.
All experiments were run on a single NVIDIA A100, H100 or H200 GPU with \texttt{torch.compile} and fp16 mixed precision, and a full run takes approximately 6 hours per seed on an A100.

All hyperparameters and sweep ranges are reported in \cref{tab:cifar100-hyperparameters}.
Across all methods, we use the same base learning rate and learning-rate schedule as \citet{dohare2024loss}.
For CBP and S\&P, we use the hyperparameters reported by \citet{dohare2024loss}.
We found that L2 init performs better when combined with usual L2 regularization (toward zero).
The remaining best-performing hyperparameters are:
\begin{itemize}
    \item \textbf{L2 init:} L2 init strength 0.0001
    \item \textbf{ReDO:} dormancy threshold 0.1, recycling period 100
    \item \textbf{Load balancing:} target range 50\%
\end{itemize}

\paragraph{Rainbow DQN.}
We use the CleanRL codebase \citep{huang2022cleanrl} for our Rainbow DQN implementation.
We did not change any of the hyperparameters.
All results are averaged over 5 random seeds.
All experiments were run on a single NVIDIA A100 or H100 GPU; a run takes about 15 hours on an A100.

\paragraph{Sensitivity of load-balancing hyperparameters.}
Load balancing introduces three hyperparameters: the target range ${[\minfiringrate, \maxfiringrate]}$, the step size $\alpha$, and the EMA decay $\avgmomentum$.
Only the target range requires tuning.

We sweep the target range over $\{ 5{-}95\%,\, 20{-}80\%,\, 40{-}60\%,\, 50\% \}$ on both supervised benchmarks, and report all four settings in \cref{tab:imagenet-accuracy,tab:cifar100-accuracy}.
On both of these backbones, accuracy increases monotonically as the band tightens: from $88.8\%$ to $89.8\%$ on continual ImageNet, and from $77.1\%$ to $78.0\%$ on class-incremental CIFAR-100, a spread of $1.0$ and $0.9$ percentage points.
The widest band is the least useful; on continual ImageNet it is indistinguishable from the BN$\,+\,$L2 reference itself ($88.8\%$), so a target range that constrains almost nothing also gains almost nothing.

The step size and the EMA decay matter far less.
On continual ImageNet, holding the BN$\,+\,$L2 backbone fixed and varying one hyperparameter at a time with two seeds per setting, we sweep $\alpha$ over $\{ 1\mathrm{e}{-04},\, 3\mathrm{e}{-04},\, 1\mathrm{e}{-03},\, 3\mathrm{e}{-03},\, 1\mathrm{e}{-02},\, 3\mathrm{e}{-02} \}$ and $\avgmomentum$ over $\{ 0.8,\, 0.9,\, 0.95,\, 0.99,\, 0.995,\, 0.999,\, 0.9995 \}$.
Across the step-size sweep, which spans a factor of $300$, accuracy over the final 100 tasks varies by $0.31$ points at a $50\%$ target and by $0.23$ points at $5{-}95\%$; across the EMA-decay sweep, by $0.35$ and $0.59$ points.
Every setting stays within $0.6$ points of the default and none of them fails, so both are dominated by the effect of the target range.
This is the basis for the defaults ${\alpha = 0.001}$ and ${\avgmomentum = 0.99}$, which we reuse unchanged across all experiments.

\paragraph{Computational performance costs.}
The load-balancing layer adds no trainable parameters.
Each layer keeps two scalars per neuron --- the averaged firing rate $\avgfiringrate$ and the shift $\beta$ --- and neither receives a gradient nor an optimizer state.
On the class-incremental CIFAR-100 ResNet-18, load balancing is installed at $17$ activations covering $3{,}904$ neurons, so it adds $7{,}808$ floats to a model with ${11{,}224{,}932}$ trainable parameters.
That is $0.07\%$ of the parameter count, and less than the BatchNorm running statistics the network already carries ($9{,}600$ floats).

At inference the layer costs a single elementwise addition, since the update in \cref{alg:load-balancing} is guarded by \texttt{self.training}.
During training it additionally computes one thresholded mean per layer, reduced over exactly the axes that BatchNorm already reduces over, and $O(d)$ work to update the two scalars.
The resulting cost is not measurable against run-to-run variation: on continual ImageNet the median wall-clock time per run on an A100 is $2.79$ hours for BN$\,+\,$L2, $2.85$ hours with load balancing at $50\%$ and $2.73$ hours at $40{-}60\%$.

\begin{table}[h]
    \centering
    \small
    \caption{\textbf{Continual ImageNet:} Hyperparameters and sweep ranges. Set-valued entries are the values considered in our sweeps.}
    \label{tab:imagenet-hyperparameters}
    \begin{tabular}{ll}
        \toprule
        \emph{Hyperparameter} & \emph{Value} \\
        \midrule
        Number of tasks            & 2000 \\
        Epochs per task            & 250 \\
        Early stopping             & No \\
        Batch size                 & 100 \\
        Optimizer                  & SGD \\
        Learning rate              & $\{ 0.001, 0.003, 0.01, 0.03, 0.1 \}$ \\
        L2 strength                & $\{ 1\mathrm{e}{-06}, 3\mathrm{e}{-06}, 1\mathrm{e}{-05}, 3\mathrm{e}{-05} \}$ or 0 \\
        Learning rate schedule     & None \\
        Momentum                   & 0.9 \\
        \midrule
        \emph{L2 init} strength    & $\{ 0.01, 0.001, 0.0001, 0.00001 \}$ \\
        \midrule
        \emph{S\&P} noise scale & $\{ 1\mathrm{e}{-07}, 1\mathrm{e}{-06}, 1\mathrm{e}{-05}, 1\mathrm{e}{-04} \}$ \\
        \midrule
        \emph{ReDO} \\
        \midrule
        Dormancy threshold         & $\{ 0, 0.01, 0.1 \}$ \\
        Recycling period           & $\{ 10, 100, 1000, 10000, 100000 \}$ \\
        \midrule
        \emph{CBP} \\
        \midrule
        Replacement rate           & $\{ 0.00003, 0.0001, 0.0003, 0.001, 0.003 \}$ \\
        Maturity threshold         & 100 \\
        Decay rate                 & 0.99 \\
        \midrule
        \emph{Load balancing} \\
        \midrule
        Target range               & $\{ 5{-}95\%, 20{-}80\%, 40{-}60\%, 50\% \}$ \\
        Step size                  & 0.001 \\
        EMA decay                  & 0.99 \\
        \bottomrule
    \end{tabular}
\end{table}

\begin{table}[h]
    \centering
    \small
    \caption{\textbf{Class-incremental CIFAR-100:} Hyperparameters and sweep ranges. Set-valued entries are the values considered in our sweeps.}
    \label{tab:cifar100-hyperparameters}
    \begin{tabular}{ll}
        \toprule
        Hyperparameter & Value \\
        \midrule
        Number of tasks            & 20 \\
        Epochs per task            & 200 \\
        Early stopping             & Yes \\
        Batch size                 & 128 \\
        Optimizer                  & SGD \\
        Base learning rate         & 0.1 \\
        LR schedule (per task)     & Step decay at epochs 60, 120, 160 \\
        Learning rate decay        & 0.2 \\
        Momentum                   & 0.9 \\
        \midrule
        \emph{L2 init} strength    & $\{ 0.01, 0.001, 0.0001, 0.00001 \}$ \\
        \midrule
        \emph{S\&P} noise scale & $1\mathrm{e}{-05}$ \\
        \midrule
        \emph{ReDO} \\
        \midrule
        Dormancy threshold         & $\{ 0, 0.01, 0.1 \}$ \\
        Recycling period           & $\{ 10, 100, 1000, 10000, 100000 \}$ \\
        \midrule
        \emph{CBP} \\
        \midrule
        Replacement rate           & $1\mathrm{e}{-05}$ \\
        Maturity threshold         & 1000 \\
        Decay rate                 & 0.99 \\
        \midrule
        \emph{Load balancing} \\
        \midrule
        Target range               & $\{ 5{-}95\%, 20{-}80\%, 40{-}60\%, 50\% \}$ \\
        Step size                  & 0.001 \\
        EMA decay                  & 0.99 \\
        \bottomrule
    \end{tabular}
\end{table}
}

\clearpage

\section{Accuracy tables}
\label{sec:accuracy-tables}

\cref{tab:imagenet-accuracy,tab:cifar100-accuracy,tab:rainbow-scores} report final test accuracies and scores for all methods on the three benchmarks, as mean $\pm$ standard error of the mean over runs.

On both supervised benchmarks the methods additionally share their randomization, which makes a stronger comparison available.
On continual ImageNet each task is a randomly sampled pair of classes, and how hard that pair is dominates the variation between runs: the BN$\,+\,$L2 reference alone spans $87.7$ to $90.0$ across its 10 runs.
That variation is a property of the task sequence, not of the method, and run $k$ of every method draws the same sequence --- so it cancels in a per-run difference, and the same $10$ differences span only $0.49$ points for LB 50\% against BN$\,+\,$L2.
On class-incremental CIFAR-100 the same holds through the seed: the class order is drawn from it, and every method is run on the same 15 seeds.

\cref{tab:imagenet-accuracy,tab:cifar100-accuracy} therefore report both quantities --- the mean with its own standard error, and the paired difference against the reference, which is resolved $2$--$5\times$ more precisely on continual ImageNet and $1.2$--$1.8\times$ on class-incremental CIFAR-100.
Comparisons between methods should be read from the paired column.

\begin{table}[h!]
\centering
\caption{\textbf{Continual ImageNet:} Final test accuracies averaged over tasks
1900--2000, with the standard error of the mean over 10 runs.
All methods are evaluated on the same 10 task sequences, so the last column reports the
\emph{paired} per-run difference against the tuned BN$\,+\,$L2 reference.
Task-sampling variance is shared and cancels there, which resolves a comparison $2$--$5\times$
more precisely than either mean for methods built on that reference.
The gain is smallest for the least stable methods --- for the Baseline the two standard
errors are the same, because its run-to-run spread is driven by its own instability rather
than by which classes it was given.
Rows are ordered by accuracy, which is also the order of the paired difference.
$^{*}\,p<0.05$, $^{**}\,p<0.001$, two-sided paired $t$-test, $\text{df}=9$.}
\label{tab:imagenet-accuracy}
\begin{tabular}{lcr@{}l}
\toprule
Method & Accuracy (\%) & \multicolumn{2}{c}{$\Delta$ vs BN$\,+\,$L2} \\
\midrule
Non-affine LN + L2 & 91.2 $\pm$ 0.22 & $+2.43 \pm 0.05$ & $^{**}$ \\
Non-affine LN + L2 + LB 50\% & 90.7 $\pm$ 0.18 & $+1.90 \pm 0.06$ & $^{**}$ \\
Non-affine BN + L2 + LB 50\% & 89.8 $\pm$ 0.18 & $+1.05 \pm 0.06$ & $^{**}$ \\
LB 50\% & 89.8 $\pm$ 0.20 & $+0.99 \pm 0.05$ & $^{**}$ \\
GELU + LB 50\% & 89.7 $\pm$ 0.19 & $+0.89 \pm 0.07$ & $^{**}$ \\
LeakyReLU + LB 50\% & 89.7 $\pm$ 0.25 & $+0.87 \pm 0.11$ & $^{**}$ \\
Non-affine BN + L2 & 89.6 $\pm$ 0.20 & $+0.85 \pm 0.07$ & $^{**}$ \\
LB 40--60\% & 89.5 $\pm$ 0.20 & $+0.72 \pm 0.04$ & $^{**}$ \\
GELU & 89.4 $\pm$ 0.20 & $+0.57 \pm 0.08$ & $^{**}$ \\
CBP @ BN + L2 & 89.3 $\pm$ 0.23 & $+0.48 \pm 0.08$ & $^{**}$ \\
LB 20--80\% & 89.1 $\pm$ 0.19 & $+0.26 \pm 0.07$ & $^{*}$ \\
CBP @ Baseline & 88.8 $\pm$ 0.24 & $+0.04 \pm 0.08$ &  \\
LB 5--95\% & 88.8 $\pm$ 0.20 & $+0.01 \pm 0.10$ &  \\
BN + L2 & 88.8 $\pm$ 0.21 & \multicolumn{2}{c}{---} \\
LeakyReLU & 88.8 $\pm$ 0.22 & $-0.01 \pm 0.05$ &  \\
ReDO @ BN + L2 & 88.7 $\pm$ 0.23 & $-0.04 \pm 0.11$ &  \\
S\&P @ BN + L2 & 88.7 $\pm$ 0.20 & $-0.12 \pm 0.08$ &  \\
L2 only & 88.6 $\pm$ 0.22 & $-0.19 \pm 0.09$ &  \\
BN only & 88.5 $\pm$ 0.22 & $-0.25 \pm 0.07$ & $^{*}$ \\
S\&P & 88.3 $\pm$ 0.24 & $-0.49 \pm 0.17$ & $^{*}$ \\
L2 init & 87.2 $\pm$ 0.23 & $-1.60 \pm 0.16$ & $^{**}$ \\
LN only & 87.2 $\pm$ 0.23 & $-1.61 \pm 0.10$ & $^{**}$ \\
LN + L2 & 87.1 $\pm$ 0.30 & $-1.69 \pm 0.13$ & $^{**}$ \\
L2 init + BN & 86.3 $\pm$ 0.22 & $-2.47 \pm 0.07$ & $^{**}$ \\
ReDO @ Baseline & 81.7 $\pm$ 0.38 & $-7.10 \pm 0.31$ & $^{**}$ \\
Baseline & 79.0 $\pm$ 0.62 & $-9.77 \pm 0.60$ & $^{**}$ \\
\bottomrule
\end{tabular}
\end{table}

\begin{table}[h!]
\centering
\caption{\textbf{Class-incremental CIFAR-100:} Final test accuracies averaged over tasks
15--20, with the standard error of the mean over 15 runs.
The class order is drawn from the run's seed and all methods use the same 15 seeds, so the
last column reports the \emph{paired} per-run difference against \emph{Resets} --- the
from-scratch reference the accuracy panels are also plotted against.
The gain from pairing is smaller here than on continual ImageNet ($1.2$--$1.8\times$ against
$2$--$5\times$): that window is 100 nearly independent tasks, this one is 5 consecutive tasks
whose difficulties are strongly correlated.
Rows are ordered by accuracy, which is also the order of the paired difference.
$^{*}\,p<0.05$, $^{**}\,p<0.001$, two-sided paired $t$-test, $\text{df}=14$.}
\label{tab:cifar100-accuracy}
\begin{tabular}{lcr@{}l}
\toprule
Method & Accuracy (\%) & \multicolumn{2}{c}{$\Delta$ vs Resets} \\
\midrule
LB 50\% & 78.0 $\pm$ 0.10 & $+1.36 \pm 0.06$ & $^{**}$ \\
LeakyReLU + LB 50\% & 77.9 $\pm$ 0.09 & $+1.24 \pm 0.05$ & $^{**}$ \\
LB 40--60\% & 77.7 $\pm$ 0.09 & $+1.05 \pm 0.06$ & $^{**}$ \\
Normalized skip connection & 77.5 $\pm$ 0.10 & $+0.84 \pm 0.05$ & $^{**}$ \\
LB 20--80\% & 77.1 $\pm$ 0.11 & $+0.51 \pm 0.06$ & $^{**}$ \\
LB 5--95\% & 77.1 $\pm$ 0.11 & $+0.43 \pm 0.06$ & $^{**}$ \\
Non-affine BN & 77.0 $\pm$ 0.10 & $+0.37 \pm 0.06$ & $^{**}$ \\
L2 init & 76.9 $\pm$ 0.08 & $+0.22 \pm 0.05$ & $^{**}$ \\
GELU + LB 50\% & 76.7 $\pm$ 0.11 & $+0.10 \pm 0.07$ &  \\
CBP & 76.7 $\pm$ 0.10 & $+0.05 \pm 0.06$ &  \\
Resets & 76.6 $\pm$ 0.09 & \multicolumn{2}{c}{---} \\
GELU & 75.9 $\pm$ 0.18 & $-0.72 \pm 0.15$ & $^{**}$ \\
LeakyReLU & 75.5 $\pm$ 0.12 & $-1.08 \pm 0.06$ & $^{**}$ \\
S\&P & 75.5 $\pm$ 0.14 & $-1.12 \pm 0.10$ & $^{**}$ \\
Baseline & 74.4 $\pm$ 0.13 & $-2.24 \pm 0.08$ & $^{**}$ \\
ReDO & 72.7 $\pm$ 0.14 & $-3.95 \pm 0.10$ & $^{**}$ \\
\bottomrule
\end{tabular}
\end{table}

\begin{table}[h!]
\centering
\revision{
\caption{\textbf{Rainbow DQN:} Final human-normalized scores ($\pm$ standard error of the mean over 5 seeds) using a time-weighted exponential moving average. LN is Rainbow DQN with LayerNorm in the hidden layers, and the last two columns are load balancing and CBP instantiated on it. Bold marks the unique best mean in each row; rows with no bold entry have a tie for the best mean.}
\label{tab:rainbow-scores}
\begin{tabular}{lcccc}
\toprule
Environment & Baseline & LN & LB 50\%$\,\text{@}\,$LN & CBP$\,\text{@}\,$LN \\
\midrule
BattleZone & 0.7 $\pm$ 0.1 & 1.0 $\pm$ 0.1 & 1.0 $\pm$ 0.1 & 0.8 $\pm$ 0.1 \\
DemonAttack & 50.6 $\pm$ 3.5 & 47.8 $\pm$ 4.5 & \textbf{55.9} $\pm$ 1.1 & 52.5 $\pm$ 1.7 \\
DoubleDunk & 8.0 $\pm$ 0.1 & 7.9 $\pm$ 0.2 & 8.0 $\pm$ 0.3 & 8.0 $\pm$ 0.2 \\
NameThisGame & 1.2 $\pm$ 0.2 & 1.8 $\pm$ 0.2 & \textbf{1.9} $\pm$ 0.1 & 1.6 $\pm$ 0.2 \\
Phoenix & 2.5 $\pm$ 0.8 & 6.3 $\pm$ 1.6 & \textbf{7.4} $\pm$ 1.3 & 3.6 $\pm$ 0.4 \\
Qbert & 1.4 $\pm$ 0.1 & 1.4 $\pm$ 0.1 & \textbf{1.6} $\pm$ 0.1 & 1.2 $\pm$ 0.1 \\
\bottomrule
\end{tabular}
}
\end{table}

\clearpage

\section{Layer-wise firing rates for all approaches}
\label{sec:layer-wise-all}

\Cref{sec:layer-wise-analysis} shows layer-wise firing rates for a representative subset of methods.
Here we give the same visualization for every approach we evaluate.

\begin{figure}
    \centerline{\includegraphics{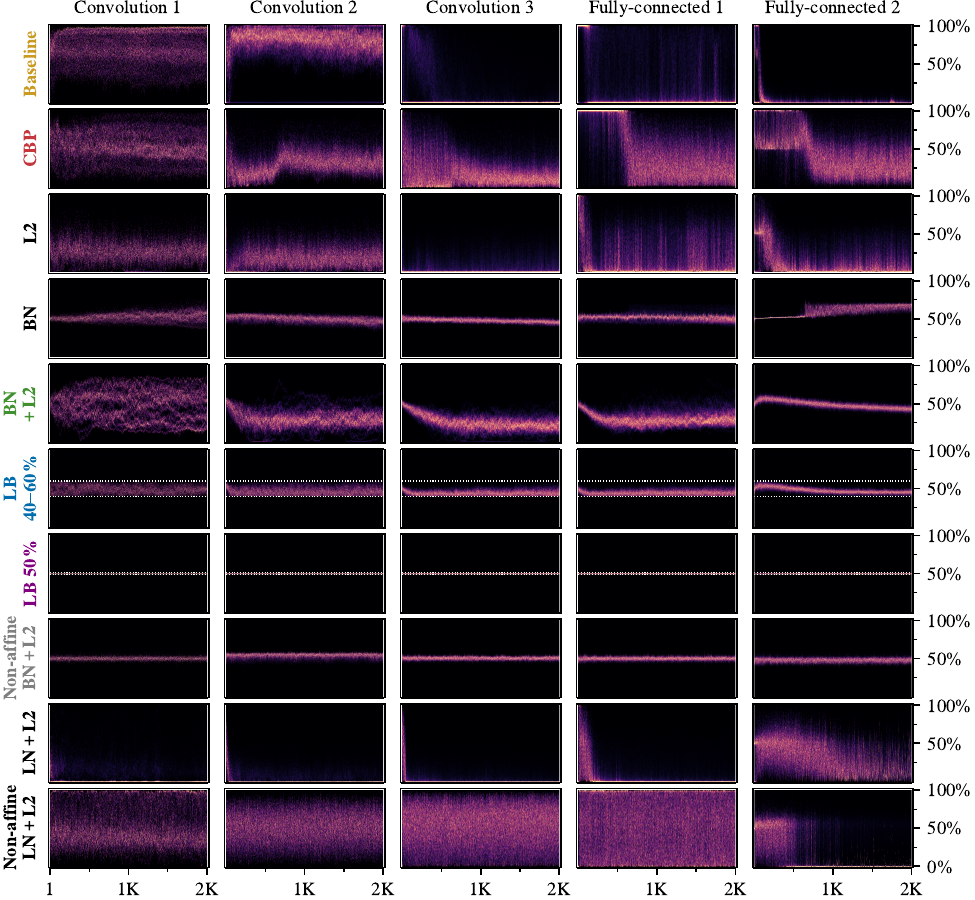}}
    \caption{\textbf{Continual ImageNet:}
    Layer-wise firing-rate distributions for all approaches.}
    \label{fig:imagenet-all-layerwise-rates}
\end{figure}

\begin{figure}
    \centerline{\includegraphics{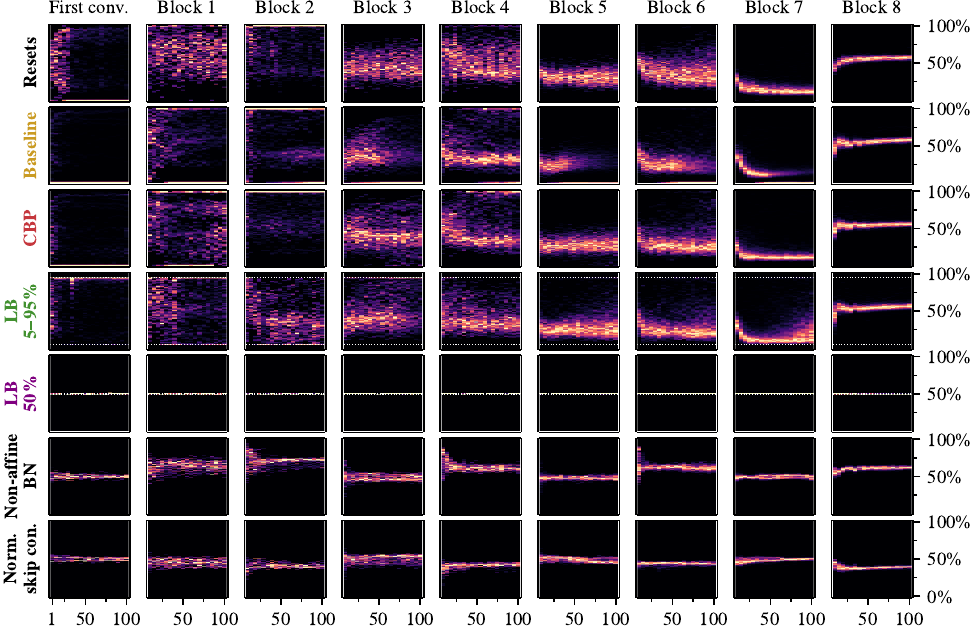}}
    \caption{\textbf{Class-incremental CIFAR-100:}
    Layer-wise firing-rate distributions for all approaches.}
    \label{fig:cifar100-all-layerwise-rates}
\end{figure}

\end{document}